\documentclass{article}
\usepackage{iclr2027_conference,times}
\usepackage{amsmath,amssymb,graphicx,booktabs,multirow,array,tabularx,enumitem,xspace,placeins,float,wrapfig}
\usepackage{hyperref}
\usepackage{url}
\usepackage{xcolor}

\definecolor{cRed}{RGB}{180, 35, 35}  
\definecolor{cBlue}{RGB}{20, 75, 160}    

\title{Locked at the Entrance, Open Inside:\\Where RLVR Narrows the Solution Space}

\author{%
  Qiancheng Zhou $^{1}$
  \quad
  Ruizhe Li$^{2\thanks{Corresponding Author: \texttt{r.li.7@bham.ac.uk}}}$
\\
  $^1$School of Future Technology, Shanghai University \\
  $^2$School of Computer Science, University of Birmingham
}

\newcommand{\task}{Countdown\xspace}
\newcommand{\passk}[1]{pass@#1\xspace}
\newcommand{\Sx}{\mathcal{S}(x)}
\newcolumntype{Y}{>{\centering\arraybackslash}X}

\iclrfinalcopy
\begin{document}
\maketitle

\begin{abstract}
Reinforcement learning with verifiable rewards (RLVR) substantially improves single-sample accuracy (\passk{1}) but causes the policy's solution space to contract, diminishing the returns of test-time scaling. In this work, we investigate where inside a reasoning trajectory this breadth is lost: does the policy fail to \textbf{access} a valid solution family, or does it fail to \textbf{execute} computation once initiated? To disentangle \textit{access} from \textit{execution}, we analyze the \task task, whose solution space can be exhaustively enumerated into discrete \textbf{entrance families} defined by the first operand and operator, across PPO on Qwen2.5-3B and GRPO on Qwen2.5-3B-Instruct. Across both training setups, solution coverage falls by up to 67\%, halving even on problems solved across all checkpoints. We show that this contraction is heavily concentrated at the entrance: per-token likelihood shifts are $11\times$--$16\times$ larger prior to the first arithmetic operation than during downstream reasoning. Supplying only an unselected entrance prefix restores completion rates in low-access families by over an order of magnitude ($0.018 \to 0.212$ under PPO), demonstrating that alternative solutions remain executable but are no longer initiated. Guided by this localization, we find that while surface prompting fails to recover diversity, entrance-targeted interventions succeed: late-layer parameter interpolation with early checkpoints increases solution coverage by 37\% at no loss in \passk{1}. Finally, we show that early-step entropy collapse recurs across six math benchmarks with 7B and 14B models, but is not an inevitable byproduct of reasoning optimization: an SFT baseline preserves more than double the coverage, and staged SFT--DPO--RLVR pipelines retain early-step entropy. In summary, reasoning breadth is lost at the door, not inside the room\footnote{Code: https://github.com/ershiyidian/early-branch-locking}.
\end{abstract}

\section{Introduction}
\label{sec:intro}

Reinforcement learning with verifiable rewards (RLVR) has become the standard paradigm for eliciting complex mathematical reasoning in LLMs~\citep{shao2024deepseekmath,deepseekai2025deepseekr1,yu2025dapo,zeng2025simplerlzoo,yan2026spurious}. In parallel, inference-time compute techniques, such as repeated sampling, self-consistency, and verifier-guided tree search, rely on generating diverse candidate trajectories to boost task accuracy \citep{brown2024largelanguagemonkeysscaling,wang2023selfconsistencyimproveschainthought,huang2025bestofnbestthemcoverage,snell2024scalingtesttime}. However, these two directions exist in fundamental tension: while RLVR substantially increases single-sample accuracy, recent studies show that post-RLVR policies collapse into narrow answer supports and exhibit severely degraded reasoning diversity~\citep{yue2025doesreinforcementlearningreally,wu2025invisibleleash,matsutani2026rlsqueezes,saha2026bodhi}, which sharply curtails the marginal gains of repeated sampling.

Existing efforts attempt to quantify this contraction or mitigate it via reward shaping, exploration bonuses, and altered rollout objectives~\citep{he2025rewardingunlikelyliftinggrpo,gai2025differentialsmoothingmitigatessharpening,li2025choicedivergenceneglectedkey,song2025outcomebasedexplorationllmreasoning}. However, the exact mechanistic issue where valid solutions are lost remains unidentified. Because prior works evaluate diversity via aggregate final-answer counts or sampled trace clustering, they conflate two distinct failure modes: a policy may fail to \textbf{access} a valid solution family (i.e., never initiating the path), or it may fail to \textbf{execute} it once accessed (i.e., derailing during downstream computation). Disentangling \textit{access} from \textit{execution} is essential, as the two failure modes dictate fundamentally different recovery mechanisms.

\begin{wrapfigure}{r}{0.42\textwidth}
\centering
\includegraphics[width=\linewidth]{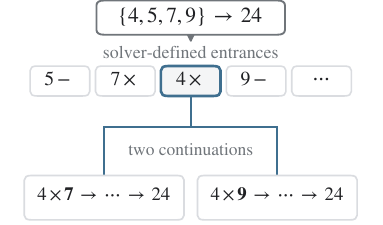}
\caption{\textbf{Entrances partition the solution set.}
For $\{4,5,7,9\}\!\to\!24$, solver-defined entrances group valid solutions by their first operand and operator.
The highlighted entrance $4\times$ admits distinct continuations beginning with $4\times7$ and $4\times9$.}
\vspace{-1em}
\label{fig:teaser}
\end{wrapfigure}

To decouple \textit{access} from \textit{execution}, we study the \task task~\citep{pan2025tinyzero}, where the entire valid solution space can be exhaustively enumerated and partitioned into discrete \textbf{entrance families} defined by the initial operand and operator (Figure~\ref{fig:teaser}). We measure free-generation coverage to quantify family \textit{access}, and probe conditional capability by clamping the prompt to a solver-specified entrance while leaving all downstream arithmetic open to measure \textit{execution}. We evaluate this framework across two distinct RLVR implementations: a PPO training trajectory on Qwen2.5-3B and an open-source GRPO checkpoint series on Qwen2.5-3B-Instruct.

\textbf{RLVR systematically induces an accuracy--breadth tradeoff across training algorithms and checkpoints} (\S\ref{sec:macro}). By tracking single-sample accuracy \passk{1} alongside exhaustive solution-space coverage across full test distribution throughout RLVR training, we observe an inverse scaling relationship: under PPO, \passk{1} increases more than fiftyfold while overall solution coverage plummets from $0.337$ to $0.111$. GRPO checkpoint series exhibits the same contraction, with accuracy tripling while solution coverage drops by $43\%$. Crucially, this contraction persists on subset of problems solved across all checkpoints, demonstrating that solution breadth collapses even when underlying problem difficulty is well within the model's capability.

\textbf{Reasoning breadth is lost at the entrance rather than through downstream execution collapse} (\S\ref{sec:mechanism}). By combining teacher-forced log-likelihood phase attribution with entrance-clamped rollouts, we isolate where probability mass shifts during training. Per-token log-likelihood divergence is $16\times$ (PPO) and $11\times$ (GRPO) larger prior to the first arithmetic operation than across all subsequent reasoning steps. Furthermore, supplying the model with a minimal, unselected entrance prefix restores completion rates in low-access families by over an order of magnitude ($0.018 \to 0.212$ under PPO), while downstream execution capability on matched prefixes strictly improves during training. Therefore, the trained policy remains capable of executing diverse solutions, and it simply stops entering them.

\textbf{Targeting opening decisions recovers solution diversity without degrading single-sample accuracy} (\S\ref{sec:interventions}). Guided by our access-localization finding, we evaluate surface-level prompts and decoding shifts against interventions that directly redistribute early computational states. While method-prompting and forced operator shifts fail to recover coverage, and temperature scaling degrades \passk{1}, entrance-aware interventions succeed: layer-specific parameter interpolation (blending late-checkpoint layers 20--28 with step-50 weights) increases solution coverage by $37\%$ at no loss in \passk{1}. Checkpoint sampling and reasoning-phase logit mixing similarly recover significant breadth by unlocking forgotten opening trajectories.

\textbf{Early entrance narrowing generalizes to multi-step math benchmarks and larger model scales} (\S\ref{sec:beyond}). By computing first-calculation entropy, same-trace likelihood profiles, and cross-checkpoint trace diversity across six standard math benchmarks (GSM8K, MATH500, Minerva Math, Olympiad-Bench, AMC23 and AIME24) on 7B and 14B Qwen architectures, we confirm that RLVR consistently triggers an early-stage distributional collapse. On extended reasoning horizons, downstream execution emerges as a compounding second bottleneck, but entrance selection remains the primary gatekeeper. Finally, evaluating alternative training paradigms reveals that breadth collapse is not an unavoidable byproduct of optimization: an SFT model maintains more than double the solution coverage, and the OLMo-3 SFT--DPO--RLVR ladder elevates \passk{1} while preserving early-step calculation entropy.

In summary, our contributions are:
\begin{enumerate}[leftmargin=*,nosep]
    \item \textbf{Access vs.\ Execution Decomposition:} We introduce an exhaustive state-space decomposition framework on \task to separate solution initiation from downstream reasoning execution.
    \item \textbf{Mechanistic Localization of Breadth Loss:} We demonstrate across PPO and GRPO that RLVR-driven diversity collapse is concentrated at the entrance ($11\times$--$16\times$ larger likelihood shift) rather than caused by downstream arithmetic failure.
    \item \textbf{Inference- and Weight-Space Recovery:} We demonstrate that parameter interpolation in late transformer layers and entrance-aware sampling recover up to $37\%$ solution coverage without sacrificing \passk{1}.
    \item \textbf{Benchmarking \& Scaling Generalization:} We confirm the entrance-narrowing phenomenon across 7B and 14B models on six standard math benchmarks, while identifying training regimes (such as staged SFT/DPO alignment) that circumvent this tradeoff.
\end{enumerate}

\section{Related Work}
\label{sec:related}

\textbf{Distributional sharpening and exploration collapse in RLVR.} Whether RLVR genuinely expands reasoning capabilities or merely reweights pretraining priors remains actively debated~\citep{liu2025prorlprolongedreinforcementlearning,wen2025reinforcementlearningverifiablerewards}. While repeated sampling scales inference-time performance~\citep{brown2024largelanguagemonkeysscaling,snell2024scalingtesttime,wang2023selfconsistencyimproveschainthought}, on-policy RLVR often triggers rapid distribution collapse, winner-take-all mode sharpening, and degraded solution coverage~\citep{mayilvahanan2025mathbeyondbenchmarkrlexpand,nguyen2025reasoningboundary,wu2025invisibleleash,yue2025doesreinforcementlearningreally,zhao2025echochamberrlposttraining}. Because updates reinforce frequently sampled successes, under-visited valid paths suffer compounding exploration penalties~\citep{yuan2026overtraining,zhou2026whenrlvrshrinks}. Recent methods counter this via modified reward functions, differential smoothing, or adaptive exploration objectives~\citep{gai2025differentialsmoothingmitigatessharpening,he2025rewardingunlikelyliftinggrpo,li2025choicedivergenceneglectedkey,song2025outcomebasedexplorationllmreasoning}. Whereas existing studies evaluate reachability at the aggregate prompt or final-answer level, we isolate where inside the reasoning trajectory valid solutions are lost, distinguishing early path initiation (\textit{access}) from downstream computation (\textit{execution}).

\textbf{Bifurcation points and prefix-guided exploration.} Prior work shows that autoregressive generation is governed by sparse, high-entropy forking tokens that disproportionately dictate downstream trajectory semantics~\citep{bigelow2024forkingpathsneuraltext,jang2026badapples,kim2026rolloutsbegin,wang20258020rulehighentropyminority}. Recent analyses track policy narrowing across semantic branches clustered from empirical rollouts~\citep{saha2026bodhi} or steer generation using reasoning prefixes~\citep{macar2026thoughtbranches,zhang2025bread}. However, sampling-based clustering cannot detect valid solution branches that the policy entirely fails to visit. By leveraging exhaustively enumerated state spaces, our framework grounds branch analysis in the ground-truth solution graph, allowing us to evaluate extinct families and probe execution capability using uninformative minimal entrance prefixes (Appendix~\ref{app:super}).

\textbf{Test-time scaling and weight-space recovery.} The efficacy of inference-time search, self-consistency, and verifier guidance is fundamentally bounded by the reachability support of the underlying policy~\citep{dragoi2025passkbreadthdepthmetricsreasoning,ju2025reasoningpathdivergencenew,yu2025passkdiagnostic}. To counteract post-training diversity loss, recent approaches ensemble temporal checkpoints or interpolate model weights across training stages~\citep{dang2025weightensembling,li2026temporalsampling}. We provide a mechanistic explanation for why these interventions succeed: parameter interpolation and checkpoint mixing act specifically by reopening dormant opening computational branches while retaining the policy's refined downstream execution capabilities (Appendix~\ref{app:related} extends this discussion).

\section{Setup: Entrances, Access, and Execution}
\label{sec:setup}

\textbf{Task formulation and training regimes.} A \task instance \citep{pan2025tinyzero} specifies a target integer and a set of three or four input operands. A valid solution is an arithmetic expression that evaluates exactly to the target using each given number precisely once with valid operations ($\{+,-,\times,\div\}$) and parentheses. To ensure algorithmic generality, we examine two independent RLVR training pipelines: \textbf{Self-Trained PPO:} Following the TinyZero framework~\citep{pan2025tinyzero}, we train Qwen2.5-3B \citep{qwen2024qwen25} with PPO, evaluating the base model alongside eleven actor checkpoints saved from step 25 to step 275. \textbf{Public GRPO Checkpoints:} We evaluate the open-source \texttt{Qwen-2.5-3B-R1-countdown} series (Qwen2.5-3B-Instruct trained with TRL GRPO). To prevent evaluation leakage, we filter the public 50{,}000-example training superset against sorted input-target semantic keys, yielding a strictly disjoint evaluation set of 135 solver-feasible problems. An independent 300-problem validation sweep selects step 25 as the earliest checkpoint achieving native-format validity above 90\%, while step 450 serves as the late endpoint. Both setups maintain their native prompt templates and formatting conventions. Because the valid state space of \task is exhaustively enumerable, we can trace the precise trajectory of disappearing solutions and monitor how probability mass migrates across the policy distribution.

\textbf{Solution space and coverage.} For any feasible instance $x$, an exact symbolic solver generates the complete set of valid arithmetic expressions, canonicalized modulo commutativity and associativity for $+$ and $\times$. We formalize this search space as a decision tree where depth-one branches represent opening arithmetic decisions and terminal leaves constitute the set of canonical valid solutions $\Sx$. Given $n$ independent model rollouts $Y_{1:n}$, the per-instance solution coverage $\mathrm{Cov}(x, Y_{1:n})$ and corpus-level mean coverage $\overline{\mathrm{Cov}}(X)$ are defined as:
\begin{equation}
\begin{aligned}
\mathrm{Cov}(x,Y_{1:n}) &= \frac{\bigl|\{s \in \Sx \mid s \text{ is the canonicalized solution of some } Y_i \in Y_{1:n}\}\bigr|}{|\Sx|}, \\
\overline{\mathrm{Cov}}(X) &= \frac{1}{|X|}\sum_{x\in X}\mathrm{Cov}(x,Y_{1:n}).
\end{aligned}
\label{eq:coverage}
\end{equation}
An independent enumerator verifies problem feasibility and leaf-set completeness on 500 held-out instances with zero discrepancy (Appendix~\ref{app:solvercheck}).

\textbf{Entrance families: Decoupling access and execution.} We stratify the initial decision space into two granularities: a coarse partition by the first arithmetic operator (\textbf{operator class}) and a fine-grained partition by the initial operand-operator tuple $(o_1, \odot_1)$, which we term the \textbf{entrance family} $b$. For instance, given the input $\{4,5,7,9\}\!\to\!24$, feasible entrance families include $5-$, $7\times$, $4\times$, and $9-$ (Figure~\ref{fig:teaser}). Across 150 held-out test problems, the solver discovers 379 feasible families (averaging 2.53 families per instance). 
Because an entrance family fixes only the initial operation while leaving all subsequent arithmetic unconstrained, solving a problem through family $b$ factorizes into an \textit{access} phase and an \textit{execution} phase:
\begin{equation}
\pi_\theta(\text{solve via } b \mid x)
= \underbrace{\pi_\theta(B=b \mid x)}_{\textit{access}}
\,\cdot\, \underbrace{\pi_\theta(\text{valid completion} \mid x, B=b)}_{\textit{execution}}.
\label{eq:factorization}
\end{equation}
We estimate family access $A_t(b\mid x)$ through unconstrained sampling and teacher-forced prefix log-probabilities. We directly measure conditional execution capability $E_t^{\mathrm{do}}(b\mid x)$ by clamping generation to a solver-constructed entrance prefix $e_b$:
$
E_t^{\mathrm{do}}(b\mid x) = \pi_t\bigl(\text{valid completion in family } b \,\big|\, x \oplus e_b\bigr)$. We use the superscript $\mathrm{do}$ to distinguish it from the observational execution term in Eq~\ref{eq:factorization}, which conditions on entrances selected by policy itself.
Throughout this paper, \emph{designated-family completion} refers to conditional probability of reaching a valid solution strictly inside the clamped family $b$, whereas \emph{any-valid success} denotes generating a valid solution in any family following the prefix.

\textbf{Supplied entrance design and information control.} To probe downstream execution without introducing confounding reasoning cues, every entrance prefix $e_b$ is synthetically constructed by the solver rather than extracted from model-generated traces (e.g., ``\texttt{Let me try: 4 *}''). Each prefix terminates immediately after the opening operator, enforcing the family constraint while leaving all downstream degrees of freedom entirely unguided. Control baselines, including neutral problem restatements and model-generated failed prefixes, verify baseline completion, while teacher-forced scoring evaluates full valid continuations. A mutual information predictability test confirms that entrance-only prefixes provide zero predictive information regarding the specific downstream leaf selection, whereas full reasoning prefixes do (Appendices~\ref{app:prefixdetail} and~\ref{app:super}).

\textbf{On-policy gradient dynamics on entrance selection.} Policy-gradient algorithms (e.g., PPO, GRPO) update policy parameters exclusively along sampled trajectories, allocating gradient updates proportional to empirical visitation frequency. Consequently, an entrance family that underperforms relative to the moving baseline during initial iterations suffers an immediate drop in selection probability. Because rarely sampled entrances receive diminishing gradient signals, initial sampling asymmetries compound over training, driving a self-reinforcing contraction of the policy's opening support \citep{yuan2026overtraining,zhou2026whenrlvrshrinks}. We formalize this via a first-order entrance surrogate in Appendix~\ref{app:pressure}, which analytically predicts that RLVR induces substantially sharper contraction in family access than in post-entrance conditional execution.

\textbf{Structural proxies for non-enumerable math benchmarks.} Because standard multi-step mathematical benchmarks lack an exhaustively enumerable solution space $\Sx$, we construct structural proxies over generated reasoning traces. A \emph{reasoning trace} is defined as the canonicalized sequence of intermediate calculation steps in a rollout, normalized by stripping formatting whitespace and sorting commutative arguments. The \emph{distinct-trace rate} measures the empirical diversity of these sequences. To capture opening diversity independent of reasoning length, we define the \emph{first calculation} as the earliest parsed token span evaluating an instantiated arithmetic relation; its categorical entropy serves as our primary metric for early-stage search breadth. Furthermore, by evaluating identical reference traces across multiple checkpoint pairs, we bifurcate per-token log-likelihoods at the boundary of the first complete calculation, directly isolating early-step distributional shifts from downstream arithmetic execution.

\textbf{Evaluation protocol and statistical rigor.} For \task, PPO evaluations are conducted using temperature $T=0.7$, top-$p=0.9$, a maximum generation budget of 256 tokens, and 320 independent rollouts across 150 held-out solver-feasible problems disjoint from the training set. GRPO evaluations use identical decoding hyperparameters ($T=0.7$, top-$p=0.9$) with a 1{,}024-token cap, 320 rollouts, and the 135 filtered instances. 
For scaling and general mathematical reasoning analyses, we evaluate Qwen2.5-7B/14B Base and SimpleRL checkpoints \citep{zeng2025simplerlzoo}, DeepSeek-R1-Distill-Qwen-7B \citep{deepseekai2025deepseekr1}, and the full OLMo-3 7B alignment progression (SFT, DPO, and RLVR) \citep{teamolmo2025olmo3}. The standard math suite spans GSM8K \citep{cobbe2021trainingverifiers} (500 problems), MATH500 \citep{hendrycks2021math,lightman2023letsverifystepstep} (500), Minerva Math \citep{lewkowycz2022solvingquantitative} (272), OlympiadBench \citep{he2024olympiadbench} (500), AMC23 (40), and AIME24 (30). Rollouts for math benchmarks are generated with 64 samples per instance at $T=0.6$, top-$p=0.95$, and a 16{,}000-token cap; truncated outputs are scored as incorrect. Accuracy is reported via the standard unbiased \passk{k} estimator \citep{chen2021evaluating}, with entropy computed in nats. All confidence intervals are estimated via bootstrap resampling (2{,}000 iterations for \task, 1{,}000 for benchmark structural statistics, and 10{,}000 for paired problem-cluster contrasts).

\section{The Accuracy--Breadth Tradeoff Across RLVR Training}
\label{sec:macro}

\textbf{Inverse scaling between accuracy and solution coverage.} Across both PPO and GRPO optimization runs, we observe a consistent inverse relationship between single-sample accuracy and the breadth of the generated solution space (Table~\ref{tab:macro}). In our self-trained PPO pipeline on Qwen2.5-3B, \passk{1} increases more than fiftyfold between steps 50 and 275, while cumulative solution coverage $\overline{\mathrm{Cov}}(X)$ at $n=320$ samples collapses by $67\%$, falling from $0.337$ to $0.111$.\footnote{Because the un-tuned base model generates $<3\%$ format-valid outputs, step 50 serves as the earliest checkpoint exhibiting stable syntactic validity for reliable breadth benchmarking.} The public GRPO series demonstrates an identical structural contraction: from step 25 to step 450, single-sample accuracy more than triples while solution coverage drops by $43\%$. This contraction directly mirrors a steep erosion in opening diversity: among valid GRPO solutions, entrance family coverage declines from $0.932$ to $0.571$, accompanied by a drop in entrance Shannon entropy from $1.585$ to $0.760$~nats (Table~\ref{tab:grpo_endpoints}). This tradeoff is invariant to decoding configurations, persisting across varying generation temperatures, top-$p$ thresholds, and maximum token budgets (Appendices~\ref{app:sboth}, \ref{app:partitions}, \ref{app:tokencap}, and~\ref{app:temperature}).

\begin{table}[t]
\centering
\caption{\textbf{Accuracy and solution breadth over two independently trained RLVR runs on \task}. PPO and GRPO use 150 and 135 held-out problems, respectively. Endpoints use 320 samples per problem and comparisons are within run. Intervals shown in Appendix~\ref{app:macroci}.}
\label{tab:macro}
\small
\begin{tabularx}{\linewidth}{lYYYYY}
\toprule
Checkpoint & \passk{1} & \passk{64} & \passk{256} & Solution cov.@320 & Operator-class cov. \\
\midrule
\multicolumn{6}{l}{\emph{PPO run}}\\
Base     & 0.001 & 0.076 & 0.228 & 0.097 & 0.178 \\
Step 25  & 0.005 & 0.218 & 0.416 & 0.189 & 0.354 \\
Step 50  & 0.051 & 0.500 & 0.596 & 0.337 & 0.538 \\
Step 75  & 0.129 & 0.482 & 0.543 & 0.283 & 0.502 \\
Step 150 & 0.323 & 0.441 & 0.457 & 0.152 & 0.362 \\
Step 275 & 0.285 & 0.376 & 0.385 & 0.111 & 0.253 \\
\midrule \multicolumn{6}{l}{\emph{Public GRPO run}}\\
Step 25  & 0.121 & 0.633 & 0.750 & 0.467 & 1.000 \\
Step 450 & 0.429 & 0.570 & 0.621 & 0.268 & 0.832 \\
\bottomrule
\end{tabularx}
\vspace{-2em}
\end{table}

\textbf{Breadth collapse persists on invariant solvable subsets.} A potential confounding hypothesis is that coverage drops simply because model forgets how to solve harder problems entirely. We rule this out by tracking solution leaf turnover on fixed subset of 58 problems solved at both early (step 50) and late (step 275) PPO checkpoints (Appendix~\ref{app:sboth}). While $64\%$ of problems solved at step 50 remain solvable at step 275, only $31\%$ of the individual valid solution leaves discovered at step 50 are ever generated by late policy (Appendix~\ref{app:leafturnover}). On this invariant solvable subset, mean coverage falls from $0.564$ to $0.286$ even as formatting compliance approaches $100\%$. Moreover, no problem missed at step 50 is newly unlocked at step 275 under 320-sample budget. Even at 2{,}048 samples per problem, 33 problems lost between steps 50 and 275 remain unrecovered (Appendix~\ref{app:sboth}). Therefore, RLVR does not broaden problem reachability; rather, it consolidates probability mass onto a narrow subset of previously accessible pathways while extinguishing alternative valid solutions.

\textbf{Contraction concentrates disproportionately at initial computational branches.} Stratifying the solution space into distinct granularities reveals that diversity loss is most severe at the earliest decision points. Among valid solutions, the coverage of coarse operator classes decreases by half over training, while coverage across the finer first-evaluated-pair partition drops by two thirds to $0.117$ (Table~\ref{tab:openpart}, Appendix~\ref{app:openpart}). Tracking next-operator distributions across pre-operator states demonstrates that entropy contraction is maximal prior to the first arithmetic operation and attenuates at downstream calculation steps. This monotonic drop in opening diversity recurs across five independent solver-enumerated structural partitions (Appendix~\ref{app:partitions}), establishing that the policy's solution space narrows primarily at its opening branch points.

\section{Mechanistic Localization: Breadth Is Lost at the Entrance}
\label{sec:mechanism}

The contraction of an RLVR policy's solution space can stem from two distinct failure modes: an \textbf{access failure} (the policy fails to initiate a valid solution branch) or an \textbf{execution failure} (the policy initiates a branch but fails to carry out the necessary downstream arithmetic). If access dominates, three empirical signatures must hold: 1). Likelihood shifts during training must concentrate predominantly around the opening arithmetic decision. 2). Supplying an unselected entrance prefix should restore completion rates in low-access families. 3). Downstream execution capability on matched supplied entrances should remain stable or improve even as autonomous entrance access collapses.
Table~\ref{tab:tests} summarizes these evaluations across both training pipelines.

\begin{table}[t]
\centering
\caption{\textbf{Access and execution diagnostics on PPO and public GRPO \task.} The first row gives early-to-late increase in NLL per token in entrance segment vs.\ subsequent execution; the second compares designated-family completion with and without a supplied minimal entrance at late checkpoint; the third tracks minimal-entrance completion between early and late checkpoints.}
\label{tab:tests}
\footnotesize
\begin{tabularx}{\linewidth}{lYY}
\toprule
Diagnostic & PPO & GRPO \\
\midrule
NLL increase per token: entrance vs.\ execution & $+4.50$ vs.\ $+0.28$ & $+8.08$ vs.\ $+0.71$ \\
Designated completion: none $\to$ minimal entrance & $0.018 \to 0.212$ & $0.104 \to 0.188$ \\
Minimal-entrance completion: early $\to$ late & $0.113 \to 0.212$ & $0.131 \to 0.188$ \\
\bottomrule
\end{tabularx}
\vspace{-2em}
\end{table}

\textbf{Likelihood shifts concentrate disproportionately prior to the first operation.} 
We isolate where probability mass migrates by computing teacher-forced negative log-likelihood (NLL) profiles over 1{,}530 paired solver-constructed valid continuations at PPO steps 25 and 275, splitting each trajectory at the boundary of the first arithmetic operator (Appendix~\ref{app:tf}). Because teacher forcing does not require the model to generate a valid completion autonomously, step 25 serves as an early, unbiased baseline. Under PPO, the entrance segment experiences a substantial likelihood drop of $4.50$~nats per token ($95\%$ CI $[4.37, 4.62]$), compared to a negligible shift of only $0.28$~nats per token ($[0.26, 0.30]$) across the remaining reasoning steps, i.e., a $16\times$ disparity in divergence. Token-level attribution reveals that the largest single likelihood drop falls directly on the initial operand token that determines family selection. On 232 valid continuations under GRPO (steps 25 vs.\ 450), the entrance shift is $8.08$~nats per token ($[7.96, 8.19]$) vs.\ $0.71$~nats ($[0.58, 0.85]$) during execution, i.e., an $11\times$ ratio. Across both algorithms, RLVR updates suppress the probability of entering alternative solution paths rather than degrading the policy's capacity to express valid downstream arithmetic.

\begin{figure}[tb]
\centering
\includegraphics[width=.98\linewidth]{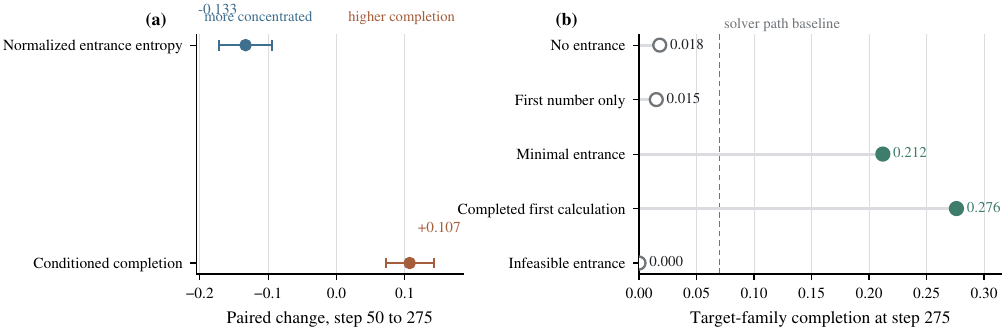}
\caption{\textbf{Access and execution move in opposite directions (PPO run).} (a)~From step 50 to step 275, normalized entrance entropy falls while completion from supplied entrances rises on matched problem and family keys; points are paired changes with 95\% problem-bootstrap intervals. (b)~At step 275, designated-family completion jumps after a minimal entrance and gains little more from the completed first calculation; the dashed line is the path-enumeration reference (Appendix~\ref{app:prefixdetail}).}
\vspace{-1.5em}
\label{fig:mechanism}
\end{figure}

\textbf{Minimal entrance prefixes restore completion in low-access families.} To test whether unvisited solution families remain executable, we prompt checkpoints with solver-constructed prefixes of varying specificity appended to a neutral scaffold: (i) no entrance, (ii) a minimal entrance specifying only the first operand and operator (e.g., ``\texttt{Let me try: 4 *}''), and (iii) a fully completed first calculation (e.g., ``\texttt{4 * 7 = 28}''). We designate up to two feasible families per problem and measure rate of valid completions falling strictly inside designated family (Table~\ref{tab:entrance}, Panel A; Figure~\ref{fig:mechanism}b). At PPO step 275, supplying a minimal entrance boosts designated-family completion by more than an order of magnitude, i.e., from $0.018$ under free sampling to $0.212$, matching the downstream efficiency of providing the fully evaluated calculation. The late GRPO checkpoint exhibits an identical recovery ($0.104 \to 0.188$). Furthermore, grafting a minimal entrance directly into the model's own failed rollouts recovers a $0.176$ valid completion rate (Panel B), whereas shorter non-arithmetic cues produce no recovery (Appendix~\ref{app:controls}). Therefore, specifying a single opening operation unlocks latent solution families without requiring external hints regarding downstream logic.

\begin{table}[t]
\centering
\caption{\textbf{Entrance interventions on the PPO \task run}. Panel A reports designated-family completion over all 150 problems, with 8 to 16 continuations per prefix and a 0.070 path-enumeration reference conditional on the entrance. In Panel B we graft the same entrances into failed traces at step 275 ($S_{\mathrm{loss}}$: 33 problems, 32 with a usable retry point; reference 0.044). Panel C covers the 28 families observed at step 50 but unobserved in 320 free samples at step 275 with $n{=}64$ continuations per cell. The GRPO counterpart is explained in Appendix~\ref{app:grpo}.}
\label{tab:entrance}
\footnotesize
\setlength{\tabcolsep}{4pt}
\renewcommand{\arraystretch}{1.06}
\begin{tabularx}{\linewidth}{@{}>{\raggedright\arraybackslash}Xccc@{}}
\toprule
& Step 50 & Step 275 & \shortstack{Step 275\\any-valid} \\
\midrule
\multicolumn{4}{@{}l}{\emph{Panel A: neutral scaffold, completion in the designated family}}\\
No entrance & 0.009 & 0.018 & 0.060 \\
Minimal entrance (operand and operator) & 0.113 & 0.212 & 0.261 \\
Completed first calculation & 0.102 & 0.276 & 0.329 \\
\midrule
\multicolumn{4}{@{}l}{\emph{Panel B: grafts into the model's own failed traces (step 275)}}\\
Feasible minimal entrance & \multicolumn{2}{>{\centering\arraybackslash}p{.44\linewidth}}{0.176 (reference 0.044; excess $+0.121$ $[0.047,0.196]$)} & 0.227 \\
Infeasible entrance & \multicolumn{2}{c}{0.000} & 0.069 \\
Empty graft & \multicolumn{2}{c}{0.000} & 0.117 \\
\midrule
\multicolumn{4}{@{}l}{\emph{Panel C: families observed at step 50 but unobserved in 320 free samples at step 275 (28 families, $n{=}64$)}}\\
All 28 families & 0.475 $[0.353,0.597]$ & 0.529 $[0.354,0.697]$ & 0.584 \\
\multicolumn{4}{@{}r}{\scriptsize paired $\Delta_{\text{275}-\text{50}} = +0.054$ $[-0.080, 0.190]$;\quad vs.\ retained families at step 275: $\Delta = -0.129$ $[-0.375, 0.126]$} \\
\bottomrule
\end{tabularx}
\vspace{-1em}
\end{table}

\textbf{Entrance access collapses while conditional execution capability improves.} We conduct a paired evaluation across 139 problems and 359 feasible family instances by matching identical $(\text{problem}, \text{family}, \text{scaffold})$ tuples between PPO steps 50 and 275. As training proceeds, entrance access concentrates sharply: opening entropy drops while top-1 family rises on matched problem and family keys. In contrast, conditional execution capability on exact same supplied entrances improves significantly, rising by $+0.107$ ($95\%$ CI $[0.073, 0.143]$; Figure~\ref{fig:mechanism}a; Appendix~\ref{app:paired}). Crucially, this preservation extends even to completely extinct pathways: for 28 entrance families observed at step 50 but never sampled at step 275 across 320 free rollouts, clamping the minimal entrance achieves a $0.529$ designated completion rate at step 275 (higher than the $0.475$ baseline at step 50; Table~\ref{tab:entrance}, Panel C; Appendix~\ref{app:unobs}). Rather than suffering internal execution collapse, alternative reasoning paths remain latent and highly competent, i.e., the trained policy simply ceases to enter them.

\section{Targeting Opening Decisions Recovers Solution Breadth}
\label{sec:interventions}

\begin{table}[tb]
\centering
\caption{\textbf{Paired intervention results on 100 held-out problems.} Each intervention uses 64 samples per problem and is compared with the same step-275 control. $\Delta$ columns report paired means with 95\% problem-cluster bootstrap intervals from 10{,}000 draws. $\dagger$ marks \passk{1} equivalence under the pre-specified $\pm0.02$ TOST margin.}
\label{tab:interventions}
\setlength{\tabcolsep}{2.5pt}
\renewcommand{\arraystretch}{1.08}
\scriptsize
\begin{tabularx}{\linewidth}{@{}>{\raggedright\arraybackslash}p{.30\linewidth}YYYY@{}}
\toprule
& \multicolumn{2}{c}{Accuracy} & \multicolumn{2}{c}{Solution breadth} \\
\cmidrule(lr){2-3} \cmidrule(lr){4-5}
Intervention & \passk{1} & $\Delta$ \passk{1} & Coverage & $\Delta$ coverage \\
\midrule
Step-275 policy (control) & 0.278 & --- & 0.111 & --- \\
Prompt diversification & $0.282^{\dagger}$ & $+0.004$ $[-0.008,0.016]$ & 0.125 & $+0.014$ $[0.000,0.038]$ \\
Forced alternative operator & 0.286 & $+0.008$ $[-0.058,0.074]$ & 0.104 & $-0.007$ $[-0.017,0.002]$ \\
Answer-only logit mixing & $0.269^{\dagger}$ & $-0.009$ $[-0.016,-0.003]$ & 0.105 & $-0.006$ $[-0.015,0.000]$ \\
High-temperature decoding ($T{=}1.0$) & 0.264 & $-0.014$ $[-0.026,-0.004]$ & 0.127 & $+0.016$ $[0.007,0.027]$ \\\hline
Reasoning-phase logit mixing & $0.277^{\dagger}$ & $-0.002$ $[-0.020,0.018]$ & 0.124 & $+0.013$ $[0.004,0.026]$ \\
Layer interpolation & 0.305 & $+0.026$ $[-0.009,0.065]$ & 0.152 & $+0.041$ $[0.020,0.069]$ \\
Checkpoint sampling & 0.161 & $-0.117$ $[-0.156,-0.081]$ & 0.211 & $+0.100$ $[0.059,0.147]$ \\
\bottomrule
\end{tabularx}
\vspace{-2em}
\end{table}

If reasoning breadth is predominantly lost at the entrance rather than during downstream execution, an immediate corollary follows: interventions that alter surface formatting or terminal answers should yield minimal recovery, whereas interventions that restore or redistribute early computational states should recover breadth in proportion to how effectively they reopen the initial entrance distribution. We evaluate this hypothesis across our saved PPO checkpoints, tuning hyperparameter configurations on a 50-problem calibration set and reporting paired evaluation results on the remaining 100 held-out problems in Table~\ref{tab:interventions} (see Appendix~\ref{app:reopen} for full protocols and ablations).

\textbf{Surface-level prompting and naive decoding adjustments fail to recover diversity.} Surface interventions applied to the late policy (step 275) produce negligible gains in solution coverage. Explicitly prompting the late policy to generate a different valid method shifts coverage by only $+0.014$ ($95\%$ CI $[0.000, 0.038]$) on paired instances. Similarly, clamping an alternative solver-feasible first operator without providing the corresponding operand state fails to improve breadth (Appendix~\ref{app:override}), and blending earlier-policy logits exclusively within final answer tags leaves coverage essentially unchanged at $0.105$. While raising decoding temperature is the sole surface-level technique that expands the generated support ($+0.016$ $[0.007, 0.027]$), it does so at the direct expense of task accuracy, i.e., degrading \passk{1} from $0.278$ to $0.264$ while recovering only a fraction of the step-50 baseline coverage (Appendix~\ref{app:reopen}). Therefore, surface manipulations cannot effectively steer the collapsed policy out of its entrenched opening modes.

\textbf{Restoring early-checkpoint representations recovers breadth without sacrificing accuracy.} In contrast, interventions that act directly on internal representations during the reasoning phase achieve substantial coverage recovery. Blending step-50 logits into non-formatting reasoning tokens increases coverage from $0.111$ to $0.124$ while maintaining late-checkpoint \passk{1}. More significantly, layer-targeted weight interpolation, i.e., linearly blending layers 20--28 of the late checkpoint with their step-50 weights, boosts solution coverage to $0.152$ at an uncompromised \passk{1} of $0.305$, representing a $37\%$ relative gain in solution breadth. At the trajectory level, splitting a fixed 64-sample generation budget evenly between steps 50 and 275 yields the largest paired coverage gain ($+0.100$ $[0.059, 0.147]$), although with single-sample accuracy shifting toward the early checkpoint. These findings provide a mechanistic grounding for prior empirical observations on temporal checkpoint sampling and weight ensembling \citep{dang2025weightensembling,li2026temporalsampling}, showing that their benefits arise specifically from unlocking dormant opening trajectories.

\textbf{Test-time allocation across feasible entrances recovers diversity without historical checkpoints.} When earlier model checkpoints or weight averages are unavailable, test-time compute can be explicitly budgeted across feasible entrance families. Under a matched per-problem token cap, distributing rollouts uniformly across solver-identified feasible entrances outperforms standard free resampling by $+0.077$ ($95\%$ CI $[0.048, 0.109]$) in solution coverage and expands the number of distinct entrance families explored by $+0.430$ ($[0.297, 0.570]$) across the 128 problems containing non-trivial failure rollouts (Appendix~\ref{app:allocation}).\footnote{The remaining 22 evaluation problems achieve a 100\% native pass rate and explore their full feasible sets.} At step 50, this structured allocation yields no significant advantage over free sampling, directly confirming our theoretical prediction: explicit entrance budgeting becomes uniquely valuable precisely when RLVR optimization causes the policy's autonomous exploration distribution to collapse.

\begin{figure}[tb]
\centering
\includegraphics[width=.98\linewidth]{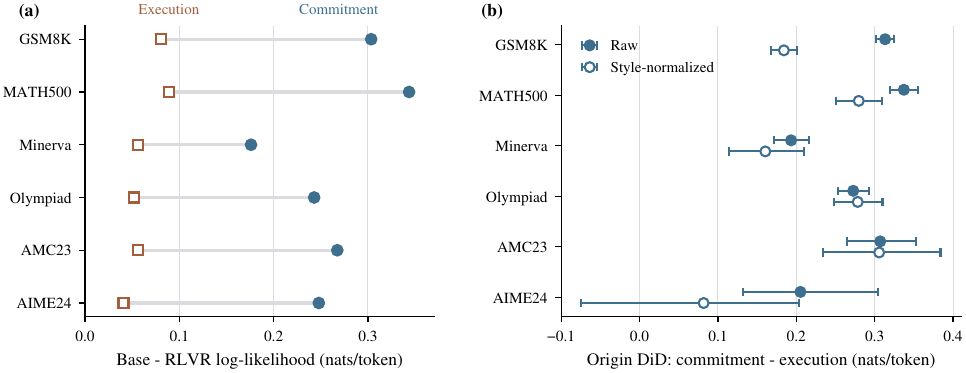}
\caption{\textbf{Early concentration on standard math benchmarks.} (a)~Same-trace scoring on Qwen2.5-7B: base minus RLVR per-token log-likelihood before the first complete calculation and during subsequent execution. (b)~Origin-stratified difference-in-differences (DiD) between the two segments, with 95\% problem-cluster bootstrap intervals (Eq.~\ref{eq:did}). Row values shown in Appendix~\ref{app:samescore}.}
\vspace{-1.5em}
\label{fig:localization}
\end{figure}

\section{Generalization Across Model Scales, Horizons, and Training Pipelines}
\label{sec:beyond}

While \task enables exact decomposition of access and execution via exhaustive state-space enumeration, we now investigate whether early entrance narrowing generalizes to multi-step math reasoning benchmarks with unconstrained solution spaces, larger parameter scales (7B and 14B), and alternative post-training pipelines.

\textbf{Early-step entropy collapse recurs across 7B and 14B model scales.} Evaluating Qwen2.5 base and SimpleRL pairs \citep{zeng2025simplerlzoo} across six standard math benchmarks demonstrates that RLVR consistently triggers early distributional collapse at scale (Table~\ref{tab:scale}; Appendix~\ref{app:scale}). While SimpleRL increases macro \passk{1} by $13\%$ at 7B and $17\%$ at 14B, asymptotic coverage remains flat, i.e., \passk{64} shifts negligibly from $0.754$ to $0.760$ at 7B and remains identical at $0.774$ at 14B. Concurrently, first-calculation entropy drops by $30\%$ at 7B and $41\%$ at 14B, while the distinct-trace rate nearly halves. Scaling the rollout budget to 256 samples per problem on GSM8K and MATH500 confirms this saturation: repeated sampling yields diminishing marginal accuracy gains under SimpleRL because early computational diversity remains several-fold lower (Appendix~\ref{app:n256}).

\textbf{Same-trace likelihood attribution confirms early-stage divergence across math benchmarks.} To localize where probability mass shifts in complex reasoning traces without alignment, we score identical reference solutions under both base and RLVR checkpoints, bifurcating per-token log-likelihoods at the boundary of the first complete calculation (Figure~\ref{fig:localization}; Appendix~\ref{app:samescore}). Across all six benchmarks, the relative likelihood divergence is strictly larger prior to the first calculation than across all subsequent reasoning steps, regardless of which policy generated the reference trace. Blinded semantic resegmentation over all 19{,}941 evaluated traces confirms that this early-segment divergence is robust across independent model-annotated boundaries (Appendix~\ref{app:semanticaudit}). Therefore, the entrance-narrowing signature identified in \task operates as the primary driver of policy concentration across standard mathematical reasoning benchmarks.

\textbf{Extended reasoning horizons introduce a compounding execution bottleneck.} In \task, selecting an entrance leaves only 6--10 tokens of downstream computation, making entrance access near-exclusive determinant of success. On open-domain math problems (GSM8K, MATH500), generating first calculation leaves hundreds of tokens of unconstrained reasoning. When we prompt RLVR policies with first-calculation prefixes that were discovered by base model but omitted under free RLVR sampling, late policy completes valid solutions at high rates ($0.850$ on GSM8K, $0.713$ on MATH500), showing that these alternative pathways remain largely executable. However, depth-controlled handoff experiments on \task reveal that as downstream reasoning length grows, execution variance increases (Appendix~\ref{app:handoff}). On complex, multi-step tasks, entrance selection remains primary gatekeeper, but long-horizon execution acts as a compounding secondary bottleneck.

\textbf{Multi-stage alignment pipelines elevate accuracy while preserving search breadth.} Solution space contraction is not an inescapable tax on reasoning capability; rather, it is sensitive to the optimization curriculum (Table~\ref{tab:posttraining}). Across the OLMo-3 7B series \citep{teamolmo2025olmo3}, \passk{1} steadily increases across the SFT $\to$ DPO $\to$ RLVR stages, but first-calculation entropy and distinct-trace rates remain remarkably stable near their initial SFT baselines \citep{karouzos2026outputdiversity}. Similarly, DeepSeek-R1-Distill-Qwen-7B \citep{deepseekai2025deepseekr1} achieves both the highest single-sample accuracy ($\text{\passk{1}}$) and the highest first-calculation entropy among all evaluated 7B models (Table~\ref{tab:deepscaler}). Applying RLVR on top of high-capacity distilled initializations can thus raise task performance while maintaining broad opening exploration distributions (Appendix~\ref{app:deepscaler}).

\begin{table}[tb]
\centering
\caption{\textbf{Post-training comparison at 7B}, using unweighted macro averages of problem means over six benchmarks; the 14B pair comparesion shown in Appendix~\ref{app:scale}. First-calculation entropy measures early-decision breadth and the distinct-trace rate is a length-sensitive complement.}
\label{tab:posttraining}
\small
\begin{tabularx}{\linewidth}{lYYYY}
\toprule
Model & \passk{1} & \passk{64} & First-calc.\ $H$ & Distinct-trace rate \\
\midrule
Qwen Base 7B     & 0.379 & 0.754 & 1.063 & 0.780 \\
Qwen SimpleRL 7B & 0.510 & 0.760 & 0.744 & 0.444 \\
Qwen Distill 7B  & 0.526 & 0.795 & 1.240 & 0.871 \\
\midrule
OLMo-3 7B (SFT)  & 0.447 & 0.781 & 1.189 & 0.770 \\
OLMo-3 7B (DPO)  & 0.508 & 0.784 & 1.269 & 0.786 \\
OLMo-3 7B (RLVR) & 0.613 & 0.828 & 1.137 & 0.763 \\
\bottomrule
\end{tabularx}
\vspace{-2em}
\end{table}

\begin{table*}[t]
\centering
\caption{\textbf{Qualitative comparison of reasoning entrances on GSM8K Item 108.} 
All models reach \passk{64}=1. Direct RLVR (\textbf{Qwen SimpleRL}) collapses onto a dominant forward equation (\textcolor{cRed}{\textbf{red}}), completely extinguishing the backward-arithmetic entrance ($10 \to 0$ out of 64 rollouts). In contrast, the staged alignment pipeline (\textbf{OLMo-3 RLVR}) preserves access to the backward deduction (\textcolor{cBlue}{\textbf{blue}}). Only initial arithmetic commitments are colored; full unedited traces shown in Appendix~\ref{app:casestudy}.}
\label{tab:qualitative_traces}
\vspace{0.4em}
\small
\setlength{\tabcolsep}{8pt}
\renewcommand{\arraystretch}{1.25}
\begin{tabularx}{\textwidth}{@{} p{0.20\textwidth} | X @{}}
\hline
\textbf{Question} & 
\textit{Henry wants to bake twice as many cookies as last year. When he finishes baking, he realizes he actually baked 15 more cookies than he meant to. He drops 5 of his cookies as he is putting them out to cool, and now has a total of 110 cookies. How many cookies did Henry bake last year? \hfill \textbf{(Gold Answer: 50)}} \\
\hline
\textbf{Qwen Base 7B} \par
{\footnotesize Sample 6; correct \par
\textcolor{cBlue}{\textbf{Reverse Entrance}}} & 
Henry now has 110 cookies. He dropped 5 cookies, so before dropping them he had \textcolor{cBlue}{\textbf{110 + 5 = 115}} cookies. He baked 15 more than he meant to, so he meant to bake 115 - 15 = 100 cookies. This is twice last year, so he baked 100 / 2 = \textbf{50} cookies last year. The answer is 50. \\
\hline
\textbf{Qwen SimpleRL 7B} \par
{\footnotesize Direct RLVR; Sample 0 \par
\textcolor{cRed}{\textbf{Forward Equation}}} & 
Let $x$ be the number of cookies Henry baked last year. He intended to bake $2x$, so he baked $2x + 15$ cookies. After dropping 5, he has \textcolor{cRed}{\textbf{(2x + 15) - 5 = 110}} cookies left. Simplifying gives $2x + 10 = 110$, so $2x = 100$ and $x = \textbf{50}$. The answer is 50. \\
\hline
\textbf{OLMo-3 7B (RLVR)} \par
{\footnotesize Staged RLVR; Sample 12 \par
\textcolor{cBlue}{\textbf{Reverse Entrance}}} & 
Henry drops 5 cookies while cooling, and then has a total of 110 cookies left. So before dropping the 5 cookies, he had: \textcolor{cBlue}{\textbf{110 + 5 = 115}} cookies. This 115 cookies is the actual number he baked: $2x + 15 = 115 \implies 2x = 115 - 15 = 100 \implies x = 100 / 2 = \textbf{50}$. The answer is 50. \\
\hline
\end{tabularx}
\vspace{-2em}
\end{table*}

\textbf{Trajectory-level case study: Direct RLVR prunes viable opening strategies.} To ground benchmark-level entropy collapse in concrete reasoning dynamics, Table~\ref{tab:qualitative_traces} details a trajectory-level case study on GSM8K Item 108, an instance where every evaluated checkpoint reaches ceiling capability (\passk{64}=1.0). The problem naturally admits three distinct reasoning entrances: an intuitive backward deduction ($110+5 \to 115-15 \to 100/2$), a forward algebraic formulation ($(2x+15)-5=110$), and a simplified net-change relation ($2x = 110-10$). While Qwen-7B Base frequently accesses the backward deduction (Sample~6; 10 of 59 correct rollouts), \textbf{Qwen SimpleRL} (direct RLVR) completely extinguishes this pathway across all 64 rollouts ($10 \to 0$), concentrating 53 completions exclusively onto the forward algebraic setup (Sample~0). In contrast, the multi-stage \textbf{OLMo-3} alignment pipeline preserves the backward entrance throughout its entire training curriculum (SFT: 1, DPO: 8, RLVR: 6), with its final RLVR checkpoint (Sample~12) retaining all three entrance formulations while achieving a perfect $64/64$ success rate. Direct RLVR thus prunes viable opening strategies even on problems within full model competence, whereas staged alignment maintains entrance plurality into late training.

\textbf{Corroboration and entrance preservation across open-domain benchmarks.} This divergence is consistent across open-domain mathematical problems. On a second GSM8K instance (Item 392, streaming discount; \passk{64}=1.0 for all models), direct RLVR concentrates 55 of 63 correct completions onto a single initial addition ($10+10=20$), causing first-step calculation entropy to collapse from $2.148$ to $0.517$~nats. Conversely, OLMo-3 RLVR sustains diverse initial operations, such as applying aggregate percentage discounts ($20 \times 0.20$) or computing itemized deductions ($10 \times 0.80$), maintaining a high first-calculation entropy of $2.429$~nats. An identical preservation pattern recurs on MATH500 (Appendix~\ref{app:math500_companion}). Because open-domain reasoning spans rely on structural parsing rather than closed-form solver graphs, full verbatim rollout sets are cataloged in Appendix~\ref{app:casestudy} for complete qualitative inspection.

\textbf{Supervised multi-solution fine-tuning retains broad solution support.} To test whether diversity preservation can be explicitly engineered, we train supervised fine-tuning (SFT) models on Qwen2.5-3B-Instruct using solver-enumerated multi-solution demonstrations across disjoint training instances, strictly holding out the 135 evaluation tasks by semantic key (Table~\ref{tab:sft_k_ablation}; Appendix~\ref{app:sft}). At $\text{\passk{1}}=0.333$, SFT maintains a solution coverage of $0.709$, i.e., more than $2.6\times$ the coverage of the late GRPO checkpoint ($0.268$ coverage at $\text{\passk{1}}=0.429$). Furthermore, sweeping the number of demonstrated solutions per problem ($k \in \{1,2,4,8\}$) reveals a strict monotonic dose--response: solution coverage scales from $0.471$ at $k=1$ to $0.766$ at $k=8$. These results confirm that solution narrowing is a specific consequence of unconstrained on-policy reinforcement dynamics rather than an inherent property of mathematical reasoning.

\section{Conclusion}
\label{sec:conclusion}

RLVR-induced solution narrowing is fundamentally an \textit{access} failure rather than an \textit{execution} collapse. Decoupling opening branch selection from downstream computation on \task reveals that probability mass drains almost entirely at the entrance ($11\times$--$16\times$ larger likelihood shifts), while latent execution capability remains intact. Consequently, surface-level prompting fails, whereas entrance-targeted interventions, such as late-layer parameter interpolation and structured entrance allocation, restore up to 38\% of solution coverage at invariant \passk{1}.
This early-step collapse recurs at 7B and 14B scales across standard math benchmarks. However, multi-solution SFT and staged alignment pipelines demonstrate that breadth loss is an artifact of unconstrained on-policy reinforcement dynamics rather than an inevitable cost of reasoning accuracy. Effective test-time scaling and training regularization must therefore target the initial decision boundary: scalable reasoning requires not only executing a chosen path to completion, but keeping the doors to alternative solutions open.



\bibliography{references}
\bibliographystyle{iclr2027_conference}

\appendix

\section{Limitations}
\label{sec:limitations}

Exhaustive enumeration is what makes access and execution separately measurable, which currently ties the direct analysis to \task. Our PPO run uses rollout group size $G{=}1$, no actor KL penalty, and a 1{,}024-token training rollout cap; the second run is a public GRPO series we evaluate rather than train; and SFT controls use LoRA fine-tuning. On standard math benchmarks the first complete calculation is an observable proxy for an entrance, not a solver-defined one. The supplied-entrance quantity $E_t^{\mathrm{do}}$ is interventional: it measures completion after an entrance is set externally, whereas the execution term in Equation~\ref{eq:factorization} conditions on entrances selected by the policy itself.

\section{Countdown Narrowing and Robustness (PPO)}
\label{app:narrowing}

This appendix collects the structural and robustness evidence behind
\S\ref{sec:macro}. The central comparisons use step 50, the earliest
format-reliable checkpoint, and step 275, the late endpoint.

\subsection{Opening partitions}
\label{app:openpart}

Table~\ref{tab:openpart} expands the early-structure measurements of
\S\ref{sec:macro}. Panel A measures access to feasible opening partitions at
two resolutions. Panel B shows where operator probability moves at matched
model-generated states immediately before an arithmetic operator, and Panel C
measures operator entropy at solver-constructed canonical prefixes along the
expression. Panel D pairs output-format validity with solution breadth.

\begin{table}[ht]
\centering
\caption{\task opening-partition diagnostics (150 problems, 320 samples per
problem). Panel B conditions on matched model-generated pre-operator
prefixes and Panel C on solver-constructed canonical prefixes.}
\label{tab:openpart}
\small
\begin{tabularx}{\linewidth}{lYY}
\toprule
& Step 50 & Step 275 \\
\midrule
\multicolumn{3}{l}{\emph{Panel A: access under two opening partitions among correct samples}}\\
Operator classes per problem (feasible) & 1.71 & 1.71 \\
Coverage of feasible operator classes & 0.538 & 0.253 \\
First-evaluated-pair labels per problem (feasible) & 3.99 & 3.99 \\
Coverage of feasible first-evaluated pairs & 0.368 & 0.117 \\
\midrule
\multicolumn{3}{l}{\emph{Panel B: operator probability at matched model-generated pre-operator prefixes}}\\
Addition & 0.812 & 0.819 \\
Subtraction & 0.088 & 0.156 \\
Division & 0.094 & 0.025 \\
Multiplication & 0.006 & 0.001 \\
Next-token operator entropy & 0.574 & 0.415 \\
\midrule
\multicolumn{3}{l}{\emph{Panel C: next-operator entropy at solver-constructed canonical prefixes (step 275)}}\\
Before op$_1$ & \multicolumn{2}{c}{1.020} \\
Before op$_2$ & \multicolumn{2}{c}{0.555} \\
Before op$_3$ & \multicolumn{2}{c}{0.479} \\
\midrule
\multicolumn{3}{l}{\emph{Panel D: output-format validity and solution breadth}}\\
Answer-parse rate & 0.240 & 0.290 \\
Tag-order validity & 0.954 & 0.997 \\
Solution coverage@320 & 0.337 & 0.111 \\
\bottomrule
\end{tabularx}
\end{table}

\subsection{On-policy pressure at entrances}
\label{app:pressure}

Let $q_b=\pi_\theta(B=b\mid x)$ be access to family $b$ and $\mu_b$ the
verifier success conditional on entering $b$ from self-generated states. With
$G$ rollouts, the probability that a training group contains a rewarded
completion through $b$ is
\begin{equation}
p_b^{+}(G)=1-(1-q_b\mu_b)^G .
\end{equation}
For $q_b\ll 1/G$, rewarded exposure scales as $Gq_b\mu_b$, and at $G{=}1$
directly as $q_b\mu_b$. For the local surrogate $J_x=\sum_b q_b\mu_b$ with
$q=\mathrm{softmax}(u)$ and fixed $\mu_b$,
\begin{equation}
\frac{\partial J_x}{\partial u_b}
=
q_b(\mu_b-\bar\mu),
\label{eq:branch_update}
\end{equation}
where $\bar\mu=\sum_j q_j\mu_j$ is mean conditional success under the current
access distribution. The gradient scales with current access, so once a
successful family becomes rare it is correspondingly less likely to supply
further positive evidence. Equation~\ref{eq:branch_update} motivates the
paired measurements of \S\ref{sec:macro}.

\subsection{Operator-class coverage over training}
\label{app:opclass}

Opening breadth declines gradually across checkpoints
(Table~\ref{tab:opclasstraj}), so the endpoint comparison is not a selection
artifact.

\begin{table}[ht]
\centering
\caption{Coverage of feasible operator classes among correct samples, base
model and all eleven RLVR checkpoints (150 problems, 320 samples).}
\label{tab:opclasstraj}
\footnotesize
\setlength{\tabcolsep}{3.2pt}
\begin{tabular}{lcccccccccccc}
\toprule
Checkpoint & Base & 25 & 50 & 75 & 100 & 125 & 150 & 175 & 200 & 225 & 250 & 275 \\
\midrule
Coverage & 0.178 & 0.354 & 0.538 & 0.502 & 0.448 & 0.412 & 0.362 & 0.328 & 0.300 & 0.262 & 0.256 & 0.253 \\
\bottomrule
\end{tabular}
\end{table}

\subsection{Leaf turnover}
\label{app:leafturnover}

For each problem solved at step 50 we track which step-50 canonical leaves
reappear later. Retention falls steadily while most of the same problems
remain solvable: by step 275, 31\% of step-50 leaves are observed again and
64\% of step-50-solved problems are still solved.

\begin{table}[ht]
\centering
\caption{Retention of step-50 correct leaves at later checkpoints, 91
step-50-solved problems, 320 samples, 2{,}000 problem-bootstrap draws.
``Zero retention'' is the fraction of problems retaining no step-50 leaf.}
\label{tab:leafturnover}
\footnotesize
\begin{tabularx}{\linewidth}{lYYY}
\toprule
Checkpoint & Leaf retention & Zero-retention fraction & Still solved \\
\midrule
Step 75  & 0.722 $[0.644,0.789]$ & 0.132 & 0.879 \\
Step 100 & 0.644 $[0.570,0.717]$ & 0.176 & 0.824 \\
Step 125 & 0.584 $[0.508,0.656]$ & 0.198 & 0.802 \\
Step 150 & 0.476 $[0.399,0.547]$ & 0.264 & 0.758 \\
Step 175 & 0.432 $[0.355,0.506]$ & 0.319 & 0.692 \\
Step 200 & 0.380 $[0.306,0.452]$ & 0.319 & 0.681 \\
Step 225 & 0.322 $[0.257,0.390]$ & 0.374 & 0.637 \\
Step 250 & 0.306 $[0.243,0.371]$ & 0.385 & 0.637 \\
Step 275 & 0.310 $[0.246,0.378]$ & 0.374 $[0.275,0.473]$ & 0.637 $[0.538,0.736]$ \\
\bottomrule
\end{tabularx}
\end{table}

\subsection{Survivorship control}
\label{app:sboth}

We split the 150 problems into $S_{\mathrm{both}}$ (58 solved at both steps 50
and 275) and $S_{\mathrm{loss}}$ (33 solved only at step 50); no problem is
solved only at step 275 under the common budget. On $S_{\mathrm{both}}$,
solution coverage falls from 0.564 to 0.286 and operator-class coverage from
0.940 to 0.654, so the contraction holds within problems that stay reachable
throughout training. On $S_{\mathrm{loss}}$ the step-275 answer-parse rate is
zero, which mixes format and access failures, so $S_{\mathrm{both}}$ carries
the within-problem breadth comparison and the supplied-entrance experiments
carry the identification. A budget stress test resamples the 33
$S_{\mathrm{loss}}$ problems and 33 matched $S_{\mathrm{both}}$ controls at
step 275 with up to 2{,}048 samples per problem: every $S_{\mathrm{loss}}$
problem remains unrecovered at every budget, while all 33 controls are
recovered within 64 samples.

\begin{table}[ht]
\centering
\caption{Survivorship control, 320 samples per problem. Coverage columns are
computed over correct samples.}
\label{tab:sboth}
\small
\begin{tabularx}{\linewidth}{llYYYYY}
\toprule
Set ($n$) & Ckpt. & \passk{1} & \passk{64} & Format-valid & Solution cov. & Operator-class cov. \\
\midrule
$S_{\mathrm{both}}$ (58) & 50  & 0.117 & 0.960 & 0.331 & 0.564 & 0.940 \\
$S_{\mathrm{both}}$ (58) & 275 & 0.737 & 0.971 & 0.749 & 0.286 & 0.654 \\
\midrule
$S_{\mathrm{loss}}$ (33) & 50  & 0.027 & 0.586 & 0.228 & 0.542 & 0.795 \\
$S_{\mathrm{loss}}$ (33) & 275 & 0.000 & 0.000 & 0.000 & 0.000 & 0.000 \\
\bottomrule
\end{tabularx}
\end{table}

\subsection{Alternative partitions}
\label{app:partitions}

Coverage falls under all six solver-enumerated partitions of the same
solution set (Table~\ref{tab:partitions}), and all endpoint intervals are
disjoint. Generation-order operator and tree-root operator are genuinely
different partitions, agreeing on only 0.218 and 0.265 of correct samples at
the two endpoints.

\begin{table}[ht]
\centering
\caption{Coverage under solver-enumerated partitions among correct samples,
150 problems, 320 samples, 2{,}000 bootstrap draws.}
\label{tab:partitions}
\footnotesize
\begin{tabularx}{\linewidth}{lYYYY}
\toprule
Partition & Feasible labels/problem & Cov.\ step 50 & Cov.\ step 275 & Relative drop \\
\midrule
Operator class (canonical) & 1.71 & 0.538 & 0.253 & 0.53 \\
Tree root operator         & 1.79 & 0.548 & 0.249 & 0.55 \\
Full operator sequence     & 3.08 & 0.424 & 0.163 & 0.62 \\
Tree signature             & 3.75 & 0.410 & 0.159 & 0.61 \\
First evaluated pair       & 3.99 & 0.368 & 0.117 & 0.68 \\
First merged numbers       & 2.49 & 0.483 & 0.167 & 0.66 \\
\bottomrule
\end{tabularx}
\end{table}

\subsection{Token-cap re-evaluation}
\label{app:tokencap}

We re-tokenize and re-evaluate all 96{,}000 endpoint samples under fixed
response caps. At the 256-token cap, pass rates match the untruncated
evaluation: 0.0508 and 0.2816 at steps 50 and 275, against 0.0513 and 0.2851.
A 128-token cap truncates many answers before the closing tags and lies below
the operating regime used here.

\subsection{Decoding sweep}
\label{app:temperature}

We sweep temperature, top-$p$, and min-$p$ over $5\times2\times2$
configurations at step 275 (Table~\ref{tab:temperature}). Higher temperature
recovers some breadth, and no setting reaches the step-50 coverage of 0.337.
The best late-policy coverage is 0.194 at $T{=}2.0$, top-$p{=}1.0$,
min-$p{=}0.05$, where \passk{1} falls to 0.238.

\begin{table}[ht]
\centering
\caption{Decoding sweep at step 275, 150 problems and 320 samples per cell.}
\label{tab:temperature}
\footnotesize
\begin{tabularx}{\linewidth}{YYYYYY}
\toprule
$T$ & top-$p$ & min-$p$ & \passk{1} & \passk{64} & Solution coverage \\
\midrule
0.7 & 0.90 & 0.00 & 0.284 & 0.377 & 0.108 \\
0.7 & 0.90 & 0.05 & 0.283 & 0.376 & 0.108 \\
0.7 & 1.00 & 0.00 & 0.277 & 0.406 & 0.138 \\
0.7 & 1.00 & 0.05 & 0.283 & 0.385 & 0.109 \\
1.0 & 0.90 & 0.00 & 0.274 & 0.402 & 0.130 \\
1.0 & 0.90 & 0.05 & 0.274 & 0.402 & 0.138 \\
1.0 & 1.00 & 0.00 & 0.267 & 0.432 & 0.149 \\
1.0 & 1.00 & 0.05 & 0.275 & 0.408 & 0.131 \\
1.3 & 0.90 & 0.00 & 0.252 & 0.425 & 0.157 \\
1.3 & 0.90 & 0.05 & 0.264 & 0.417 & 0.151 \\
1.3 & 1.00 & 0.00 & 0.222 & 0.445 & 0.169 \\
1.3 & 1.00 & 0.05 & 0.267 & 0.434 & 0.149 \\
1.6 & 0.90 & 0.00 & 0.154 & 0.438 & 0.158 \\
1.6 & 0.90 & 0.05 & 0.260 & 0.438 & 0.170 \\
1.6 & 1.00 & 0.00 & 0.107 & 0.441 & 0.184 \\
1.6 & 1.00 & 0.05 & 0.256 & 0.448 & 0.165 \\
2.0 & 0.90 & 0.00 & 0.034 & 0.401 & 0.144 \\
2.0 & 0.90 & 0.05 & 0.239 & 0.446 & 0.174 \\
2.0 & 1.00 & 0.00 & 0.009 & 0.295 & 0.147 \\
2.0 & 1.00 & 0.05 & 0.238 & 0.467 & 0.194 \\
\bottomrule
\end{tabularx}
\end{table}

\subsection{Bootstrap intervals}
\label{app:macroci}

Table~\ref{tab:macroci} provides the complete trajectory of point estimates and 95\% confidence intervals for the PPO \task run summarized in Table~\ref{tab:macro}, evaluated across all 150 held-out problems with 320 rollouts per instance. All confidence intervals are computed via problem-cluster bootstrap with 2{,}000 resample draws (seed 1729) to account for problem-level variance. Crucially, the 95\% confidence intervals for solution coverage at the earliest format-competent checkpoint (Step~50: $[0.277, 0.397]$) and the late checkpoint (Step~275: $[0.079, 0.145]$) are strictly disjoint, confirming that the two-thirds collapse in solution space breadth is statistically significant and not an artifact of rollout sampling noise. In parallel, while single-sample accuracy (\passk{1}) increases monotonically across non-overlapping intervals, multi-sample capability (\passk{64}) plateaus and gradually deteriorates, reinforcing the structural accuracy--breadth tradeoff characterized in \S\ref{sec:macro}.

\begin{table}[ht]
\centering
\caption{Problem-bootstrap intervals for the PPO \task run, 150 problems and
320 samples per problem, 2{,}000 draws, seed 1729.}
\label{tab:macroci}
\footnotesize
\begin{tabularx}{\linewidth}{lYYY}
\toprule
Checkpoint & \passk{1} & \passk{64} & Solution coverage@320 \\
\midrule
Base     & 0.001 $[0.001,0.002]$ & 0.076 $[0.054,0.099]$ & 0.097 $[0.063,0.135]$ \\
Step 25  & 0.005 $[0.004,0.006]$ & 0.218 $[0.174,0.263]$ & 0.189 $[0.145,0.235]$ \\
Step 50  & 0.051 $[0.040,0.062]$ & 0.500 $[0.431,0.579]$ & 0.337 $[0.277,0.397]$ \\
Step 75  & 0.129 $[0.100,0.159]$ & 0.482 $[0.407,0.561]$ & 0.283 $[0.230,0.337]$ \\
Step 100 & 0.178 $[0.134,0.224]$ & 0.446 $[0.373,0.525]$ & 0.229 $[0.181,0.278]$ \\
Step 125 & 0.276 $[0.214,0.340]$ & 0.446 $[0.373,0.526]$ & 0.205 $[0.156,0.254]$ \\
Step 150 & 0.323 $[0.256,0.393]$ & 0.441 $[0.365,0.521]$ & 0.152 $[0.116,0.190]$ \\
Step 175 & 0.288 $[0.223,0.357]$ & 0.399 $[0.324,0.480]$ & 0.146 $[0.109,0.187]$ \\
Step 200 & 0.297 $[0.231,0.366]$ & 0.401 $[0.327,0.480]$ & 0.122 $[0.088,0.160]$ \\
Step 225 & 0.247 $[0.184,0.312]$ & 0.382 $[0.309,0.459]$ & 0.113 $[0.081,0.148]$ \\
Step 250 & 0.247 $[0.186,0.311]$ & 0.381 $[0.307,0.458]$ & 0.110 $[0.079,0.144]$ \\
Step 275 & 0.285 $[0.217,0.351]$ & 0.376 $[0.303,0.453]$ & 0.111 $[0.079,0.145]$ \\
\bottomrule
\end{tabularx}
\end{table}

\subsection{Solver cross-check}
\label{app:solvercheck}

A second enumerator shares no code with the primary solver, using ordered
operand pairs and tuple-tree serialization instead of abstract-syntax-tree
canonicalization. The two implementations agree on feasibility and on the
complete canonical leaf set for all 500 held-out instances.

\section{Entrance Protocol and Evidence (PPO)}
\label{app:entrance}

This appendix gives the construction, controls, and matched analyses behind
\S\ref{sec:mechanism}.

\subsection{Prefix construction and predictability gate}
\label{app:prefixdetail}

Each instance combines a scaffold with a solver-constructed prefix. The
\emph{neutral} scaffold restates the numbers and target in the model's normal
format, and the \emph{retry} scaffold places the prefix after an unsuccessful
attempt. The conditions are:
\begin{itemize}[leftmargin=*,nosep]
\item \textbf{No entrance}: no arithmetic entrance is supplied;
\item \textbf{Minimal entrance}: the designated family's first operand and
operator after ``\texttt{Let me try:}'';
\item \textbf{Completed first calculation}: the first calculation with its
intermediate value.
\end{itemize}

Designated families are taken from the feasible set in decreasing family
size, up to two per problem; of the 150 problems, 45 contribute one family and
105 contribute two. Family membership is scored from the final
\texttt{<answer>} expression by the canonical parser, which reproduces the
declared family for all 359 solver-constructed examples, with 30 continuations
additionally checked by hand. Each prefix receives 8 to 16 continuations at
temperature 0.7.

Three reference quantities differ in role. The \emph{path-enumeration
reference}, 0.070 on average, is the fraction of enumerated arithmetic paths
that reach the target once the designated operand and operator are fixed. The
\emph{leaf-uniform family share}, 0.396, is the fraction of canonical leaves
belonging to the designated family. Natural access to that family at step 275
is 0.018.

The predictability gate scores whether a prefix reveals downstream solution
information beyond family identity, by asking whether the remaining
expression can be read off the prefix. Entrance-only and empty prefixes pass
on all 150 problems with median score 0; full successful-reasoning prefixes
fail on all 150 with median score 1.

On the four-number intersection of 82 problems, step-275 minimal-entrance
completion is 0.049 against a 0.013 path-enumeration reference, and 0.015 at
step 50. Holding problem, family, and retry scaffold fixed, completion on this
subset rises from 0.012 at step 25 to 0.044 at step 100 and 0.093 at step 200,
so the effect is not specific to one late checkpoint. Six semantically
equivalent minimal-entrance templates yield designated completion between
0.269 and 0.350, so the recovery effect does not depend on the surface wording
of the cue.

\subsection{Intermediate prefixes and controls}
\label{app:controls}

To find the smallest useful prefix we vary what is added at the retry point
(Table~\ref{tab:controls}). Generic retry text, the first number alone, and
generic or misleading plans stay at baseline, while a completed local
calculation produces a large effect. The useful intervention therefore sits
between a bare cue and a downstream plan: it fixes one concrete arithmetic
action and leaves the rest open.

\begin{table}[ht]
\centering
\caption{Intermediate conditions on the retry scaffold, designated-family
completion over 139 problems.}
\label{tab:controls}
\footnotesize
\begin{tabularx}{\linewidth}{lYYYYY}
\toprule
Checkpoint & Retry cue & First number & Local calc. & Generic plan & Misleading plan \\
\midrule
Step 50  & 0.031 & 0.037 & 0.147 & 0.024 & 0.040 \\
Step 275 & 0.015 & 0.015 & 0.211 & 0.023 & 0.004 \\
\bottomrule
\end{tabularx}
\end{table}

\subsection{Teacher-forced likelihood}
\label{app:tf}

For each instance we render a solver-constructed valid continuation and
teacher-force the same continuation at steps 25 and 275, splitting the target
at its first arithmetic operator. Over 1{,}530 paired states the entrance
segment loses $4.496$ nats per token $[4.371,4.623]$, against $0.280$
$[0.263,0.297]$ during subsequent execution. In total NLL the changes are
$8.504$ $[8.280,8.729]$ and $3.063$ $[2.884,3.247]$. Teacher forcing needs no
well-formatted generated answer, so step 25 can be scored even though step 50
is the free-sampling reference.

Aligning the same continuations at the first arithmetic operator gives the
per-token profile in Table~\ref{tab:tokenprofile}. The largest increase falls
on the operand token that selects the family, immediately before the first
operation.

\begin{table}[ht]
\centering
\caption{Teacher-forced per-token negative log-likelihood by position relative
to the first arithmetic operator. Position 0 is the first token at or after the
boundary. Intervals use problem-cluster bootstrap with 10{,}000 draws.}
\label{tab:tokenprofile}
\footnotesize
\begin{tabularx}{\linewidth}{rYYYr}
\toprule
Relative position & Step 25 & Step 275 & $\Delta$ & Tokens \\
\midrule
$-4$ & 3.643 $[3.301,3.977]$ & 5.531 $[4.798,6.242]$ & $+1.888$ & 264 \\
$-3$ & 8.602 $[7.975,9.182]$ & 16.108 $[14.955,17.232]$ & $+7.506$ & 1404 \\
$-2$ & 2.098 $[1.779,2.448]$ & 3.315 $[2.715,3.979]$ & $+1.217$ & 1530 \\
$-1$ & 2.428 $[2.200,2.657]$ & 3.358 $[2.978,3.760]$ & $+0.930$ & 1530 \\
$0$  & 0.982 $[0.824,1.153]$ & 1.304 $[1.022,1.602]$ & $+0.321$ & 1530 \\
$+1$ & 0.762 $[0.668,0.863]$ & 1.230 $[1.050,1.420]$ & $+0.468$ & 1530 \\
$+2$ & 0.923 $[0.709,1.155]$ & 1.113 $[0.847,1.401]$ & $+0.190$ & 1530 \\
\bottomrule
\end{tabularx}
\end{table}

\subsection{Failed-trace grafts}
\label{app:graft}

To test the entrance inside model-generated context we build grafts from
step-275 traces that have already made a verified unsuccessful attempt. At the
retry boundary we append a solver entrance for a feasible family, a
solver-verified infeasible entrance, or no arithmetic state.

On $S_{\mathrm{loss}}$, 32 of 33 problems admit a usable retry point. The
feasible condition holds 84 graft instances: designated-family completion is
0.176 $[0.093,0.263]$, any-valid success is 0.227 $[0.122,0.338]$, and the
excess over the 0.044 path-enumeration reference is $+0.121$ $[0.047,0.196]$.
The infeasible condition holds 64 instances with zero designated completion
and 0.069 any-valid success, and the empty graft gives zero designated
completion with 0.117 any-valid. A neutral-scaffold replication on
$S_{\mathrm{loss}}$ yields excess 0.062 $[0.006,0.132]$. The minimal entrance
therefore reopens a feasible family inside the late policy's own failed
reasoning, not only from a clean prompt.

\subsection{Paired access and execution}
\label{app:paired}

Matching steps 50 and 275 on identical (problem, family, scaffold) keys gives
139 problems and 359 feasible family instances. Because per-family access
shares sum to one within the designated set, access concentration is
summarized at the set level. From step 50 to step 275, problem-paired entrance
entropy falls by $0.144$ $[0.104,0.186]$ and normalized entropy by $0.133$
$[0.095,0.172]$, top-1 entrance share rises by $0.074$ $[0.051,0.098]$, and
Gini rises by $0.062$ $[0.042,0.082]$. Mean absolute per-family access change
is 0.115 $[0.098,0.134]$.

On the same supplied entrances, designated-family completion improves by
$+0.107$ $[0.073,0.143]$. Among instances whose step-275 access is below 0.05,
mean completion is 0.216 against a mean path-enumeration reference of 0.054;
for the rest it is 0.313 against 0.068. This pairing is the central contrast:
training concentrates access over entrance families while improving completion
from matched supplied entrances.

The free-generation traces also permit an observational decomposition of the
change in solve probability. For each feasible family we estimate the
execution term of Equation~\ref{eq:factorization} from rollouts that enter the
family without intervention and apply the exact two-factor Shapley identity to
the change between steps 50 and 275. Aggregated over the 150 problems, the
decomposition assigns 34.5\% of the change in solve probability to access
reallocation and 65.5\% to conditional execution, exact up to machine
precision. Access contracts the support over which solutions are sampled,
while improved conditional execution supplies most of the increase in solve
probability.

\subsection{Access-tail recoverability}
\label{app:unobs}

We call an entrance family \emph{unobserved} when its access probability is at
least $0.05$ at step 50 and zero across 320 free samples at step 275 (28
families), and \emph{retained} when access remains at least $0.05$ at both
checkpoints (24 families). Supplying the minimal entrance to the unobserved
families yields designated-family completion 0.529 $[0.354,0.697]$ at step 275
and 0.475 $[0.353,0.597]$ at step 50, well above their mean path-enumeration
reference of 0.115; retained families reach 0.658 $[0.464,0.831]$ at step 275.

Recoverability varies across the access tail. On a separate 800-problem
train-disjoint pool, a broader set of 276 zero-access family cells spanning
250 problems achieves mean supplied-entrance completion 0.087
$[0.057,0.120]$. Conditional recoverability is therefore graded rather than
binary: branches that once carried appreciable policy mass reopen readily,
while the deeper tail is harder to reach from the same cue.

The zero-access designation itself is stable under larger budgets. On 150
held-out problems we draw 1{,}024 late-policy samples and split them into two
independent halves. Of 337 families absent from the first 512 samples, 336
remain absent from the second 512, none exceeds 0.02 access in the held-out
half, and a zero hit gives a 95\% Clopper--Pearson upper bound of 0.0058.

\subsection{Successful-trace prefixes}
\label{app:super}

An alternative is to cut prefixes from the model's own successful reasoning
just before or after its first answer operator. These produce high downstream
completion and also carry the remaining plan. At step 275 they yield
completion 0.617 before the first operator and 0.999 after it, and full
successful reasoning fails the predictability gate on all 150 problems even
with no final expression present. Lexical screening does not remove that
information, so the identification analysis uses solver-constructed entrances.
Table~\ref{tab:recovery_gradient} contrasts the two.

\begin{table}[ht]
\centering
\caption{Successful-trace prefixes and gate-passing controls. Panel A is a
state-only ablation at step 200 on 29 problems with 16 continuations per
condition. Panel B uses 184 successful-trace boundary instances with 64
continuations. Panel C gives solver-constructed controls. The successful
natural-language and successful-trace conditions fail the predictability gate;
the others pass.}
\label{tab:recovery_gradient}
\scriptsize
\setlength{\tabcolsep}{4pt}
\begin{tabularx}{\linewidth}{>{\raggedright\arraybackslash}XYYYY}
\toprule
Condition & Ckpt. & Problems $\times$ cont. & Prefix tokens & Rate \\
\midrule
\multicolumn{5}{l}{\emph{Panel A: state-only ablation}}\\
Numeric multiset only & 200 & $29\times16$ & 12.9 & 0.002 \\
Multiset and first operator (bare label) & 200 & $29\times16$ & 14.9 & 0.017 \\
Successful natural-language scaffold & 200 & $29\times16$ & 171.1 & 0.998 \\
\midrule
\multicolumn{5}{l}{\emph{Panel B: successful-trace boundary contrast}}\\
Before op$_1$ & 50 & $184\times64$ & --- & 0.145 \\
After op$_1$  & 50 & $184\times64$ & --- & 0.914 \\
Before op$_1$ & 275 & $184\times64$ & --- & 0.617 \\
After op$_1$  & 275 & $184\times64$ & --- & 0.999 \\
\midrule
\multicolumn{5}{l}{\emph{Panel C: solver-constructed controls}}\\
Solver entrance after op$_1$, low-access family & 275 & solver set & --- & 0.133 \\
Solver entrance after op$_1$, dominant family & 275 & solver set & --- & 0.193 \\
Random syntactic after op$_1$ (any-valid) & 275 & solver set & --- & 0.082 \\
Dead-end full expression (valid / format-closing) & 275 & $20\times8$ & --- & 0.000 / 1.000 \\
\bottomrule
\end{tabularx}
\end{table}

\section{Replication on a Public GRPO Run}
\label{app:grpo}

\paragraph{Released checkpoints and protocol.}
The second \task setting uses the public checkpoint series
\texttt{philschmid/qwen-2.5-3b-r1-countdown}, based on Qwen2.5-3B-Instruct and
trained with TRL GRPO on \texttt{Jiayi-Pan/Countdown-Tasks-3to4}. We evaluate
the released checkpoints in their native prompt and output format, including
the assistant prefill that begins generation inside \texttt{<think>}.

\paragraph{Overlap-excluded evaluation set.}
The public data preparation shuffles the source dataset with seed 42 and takes
the first 50{,}000 examples before its train and test split. Since the exact
split cannot be reconstructed from the released code, we treat all 50{,}000 as
a training superset, define the semantic key of a problem as its sorted input
numbers with the target, and remove every match from our 150-problem set.
Fifteen problems overlap, leaving 135 for all GRPO analyses.

\paragraph{Checkpoint selection.}
Checkpoint selection uses an independent 300-problem validation set disjoint
from the 135-problem evaluation set by semantic key. We evaluate all eight
released checkpoints with 64 samples per problem at temperature 0.7,
top-$p=0.9$, and a 1{,}024-token cap, and select the earliest checkpoint whose
native-format rate exceeds 0.90. This rule selects step 25; step 450 is the
late endpoint. Coverage declines steadily along the validation trajectory
(Table~\ref{tab:grpo_curve}), with a problem-level regression on
$\log(\mathrm{step})$ giving slope $-0.117$ (95\% CI $[-0.134,-0.099]$).

\begin{table}[ht]
\centering
\caption{Independent validation trajectory for the public GRPO run (300
problems, 64 samples per problem). The evaluation set used for endpoint
comparisons is not used for checkpoint selection.}
\label{tab:grpo_curve}
\footnotesize
\begin{tabularx}{\linewidth}{lYYYYY}
\toprule
Checkpoint & \passk{1} & \passk{64} & Native-format & Solution cov. & Entrance cov. \\
\midrule
Step 25  & 0.142 & 0.703 & 0.905 & 0.358 & 0.766 \\
Step 50  & 0.181 & 0.710 & 0.947 & 0.384 & 0.735 \\
Step 100 & 0.321 & 0.700 & 0.903 & 0.385 & 0.637 \\
Step 150 & 0.450 & 0.657 & 0.864 & 0.304 & 0.540 \\
Step 200 & 0.489 & 0.670 & 0.941 & 0.279 & 0.479 \\
Step 300 & 0.507 & 0.673 & 0.947 & 0.267 & 0.479 \\
Step 400 & 0.515 & 0.660 & 0.956 & 0.254 & 0.471 \\
Step 450 & 0.516 & 0.663 & 0.955 & 0.258 & 0.473 \\
\bottomrule
\end{tabularx}
\end{table}

\subsection{Breadth at the selected endpoints}

At 320 samples per problem, single-sample accuracy rises sharply while
solution coverage, entrance coverage, and entrance entropy all narrow
(Table~\ref{tab:grpo_endpoints}). The supplemental reasoning-derived
partition sharpens further, with think-entrance entropy falling from 2.225 to
0.081. We report the answer-defined partition as primary because it aligns
with the solver-enumerated solution families used throughout.

\begin{table}[ht]
\centering
\caption{Endpoint evaluation for the public GRPO run (135 problems, 320
samples per problem).}
\label{tab:grpo_endpoints}
\footnotesize
\begin{tabularx}{\linewidth}{lYYYYYYY}
\toprule
Checkpoint & \passk{1} & \passk{64} & \passk{256} & Native-format rate & Solution cov. &
Entrance cov. & Entrance $H$ \\
\midrule
Step 25  & 0.121 & 0.633 & 0.750 & 0.900 & 0.467 & 0.932 & 1.585 \\
Step 450 & 0.429 & 0.570 & 0.621 & 0.941 & 0.268 & 0.571 & 0.760 \\
\bottomrule
\end{tabularx}
\end{table}

\subsection{Teacher-forced likelihood}

Scoring the same 232 solver-constructed valid continuations at both selected
checkpoints and splitting NLL at the first arithmetic operator, the increase
is far larger at the entrance than during execution
(Table~\ref{tab:grpo_tf}). The per-token entrance shift is about eleven times
the execution shift, reproducing the asymmetry under a different RLVR recipe
and prompt format.

\begin{table}[ht]
\centering
\caption{Teacher-forced likelihood on the public GRPO run. Values are NLL per
token with 95\% problem-cluster bootstrap intervals.}
\label{tab:grpo_tf}
\small
\begin{tabularx}{\linewidth}{lYY}
\toprule
Checkpoint or contrast & Entrance & Execution \\
\midrule
Step 25  & 6.840 $[6.741,6.943]$ & 1.078 $[0.995,1.162]$ \\
Step 450 & 14.917 $[14.731,15.119]$ & 1.788 $[1.593,1.984]$ \\
Late $-$ early & $+8.077$ $[7.962,8.193]$ & $+0.709$ $[0.578,0.846]$ \\
\bottomrule
\end{tabularx}
\end{table}

\subsection{Supplied entrances}

We repeat the entrance progression in the released model's native prompt, with
16 continuations per condition and problem (Table~\ref{tab:grpo_staircase}).
At step 450, naming only the opening operand and operator raises
designated-family completion from 0.104 to 0.188, and the completed first
calculation reaches 0.225. Minimal-entrance completion also rises over
training, from 0.131 to 0.188. As in the PPO run, access concentrates while
execution from a supplied entrance stays available and improves.

\begin{table}[ht]
\centering
\caption{Supplied entrances on the public GRPO run. The primary quantity is
valid completion in the designated entrance family.}
\label{tab:grpo_staircase}
\small
\begin{tabularx}{\linewidth}{llYYY}
\toprule
Checkpoint & Prefix & Designated completion & Any-valid & Native-format rate \\
\midrule
Step 25  & none & 0.056 & 0.116 & 0.911 \\
Step 25  & minimal & 0.131 & 0.211 & 0.743 \\
Step 25  & first calc. & 0.143 & 0.198 & 0.843 \\
\midrule
Step 450 & none & 0.104 & 0.434 & 0.941 \\
Step 450 & minimal & 0.188 & 0.354 & 0.793 \\
Step 450 & first calc. & 0.225 & 0.335 & 0.778 \\
\bottomrule
\end{tabularx}
\end{table}

\section{Reopening Entrances (PPO)}
\label{app:reopen}

The intervention comparison uses a fixed 50-problem tuning split and a
100-problem confirmation split of the 150 PPO evaluation problems.
Hyperparameters are selected on the tuning split and frozen before
confirmation. Every confirmation arm uses 64 samples per problem with matched
problem identities and the same step-275 control; intervals use 10{,}000
problem-cluster bootstrap draws, and \passk{1} equivalence is assessed by
paired two one-sided tests at a pre-specified margin of $\pm0.02$.
Table~\ref{tab:interventions} reports the confirmation estimates.

\paragraph{Intervention definitions.}
Prompt diversification adds a request for a different valid method. The forced
operator arm constrains the opening operator without supplying an arithmetic
state (Appendix~\ref{app:override}). Answer-only and reasoning-phase logit
mixing interpolate step-50 logits either inside the answer span or during
reasoning, respectively. High-temperature decoding raises the sampling
temperature from 0.7 to 1.0. Layer interpolation replaces blocks 20--28 by the
midpoint of their step-50 and step-275 parameters. Checkpoint sampling
allocates 32 of the 64 rollouts to each endpoint checkpoint.

\paragraph{Confirmation pattern.}
Prompt diversification, answer-only mixing, and reasoning-phase mixing satisfy
the $\pm0.02$ \passk{1} equivalence criterion; of these, reasoning-phase
mixing produces the clearest breadth gain, $+0.013$ $[0.004,0.026]$. Layer
interpolation gives the largest single-policy gain, $+0.041$ $[0.020,0.069]$,
with observed \passk{1} increasing from 0.278 to 0.305. Checkpoint sampling
raises coverage by $+0.100$ $[0.059,0.147]$ while moving single-sample
accuracy toward the early checkpoint. The confirmation results match the
ordering predicted by the localization: interventions that redistribute early
reasoning states recover breadth, while surface edits do not.

\subsection{Forced operator override}
\label{app:override}

Starting from successful step-275 samples, the forced-operator arm identifies
an alternative first operator that can open at least one valid expression
under the same first number, without supplying an arithmetic state. On the
confirmation split it changes solution coverage by $-0.007$
$[-0.017,0.002]$. Selecting a token is not supplying the local state that
identifies a feasible branch; the latter is the effective entrance
intervention of \S\ref{sec:mechanism}.

\subsection{Cross-run confirmation on public GRPO}

We evaluate representative early-state interventions on the independent public
GRPO trajectory under the same confirmation protocol. Answer-only mixing
preserves \passk{1} within the $\pm0.02$ equivalence margin
($\Delta=-0.0006$, 95\% CI $[-0.0119,0.0086]$) while increasing
first-operation family coverage by $+0.030$ $[0.001,0.065]$. Checkpoint
sampling again recovers the largest breadth gain (first-operation coverage
$+0.132$) while moving \passk{1} toward the early checkpoint. The same
breadth--accuracy ordering reappears on an independently trained run with a
different RLVR algorithm and prompt format.

\subsection{Entrance allocation}
\label{app:allocation}

For each problem we take up to three failed samples as scaffolds, append each
feasible solver entrance, and sample 16 continuations per entrance. Budget
matching uses one per-problem token cap, the minimum of the full 320-sample
generated-token budget and the full allocation cost, and both conditions are
truncated deterministically to it.

At step 275, 128 problems have complete outputs in both conditions. There,
entrance allocation exceeds same-problem free resampling by $+0.148$
$[0.086,0.211]$ in any-valid success, $+0.077$ $[0.048,0.109]$ in coverage,
and $+0.430$ $[0.297,0.570]$ in distinct entrance families. At step 50 all 150
problems are comparable and the contrasts are flat: any-valid $-0.040$
$[-0.107,0.027]$ and coverage $-0.001$ $[-0.044,0.042]$. Allocating attempts
across feasible entrances therefore helps precisely once access has
concentrated.

\section{Math Benchmarks: Early Concentration and Its Limits}
\label{app:transfer}

Exhaustive enumeration is unavailable here, so the transfer test uses
observable structure: diversity of first calculations, distinct traces, and
same-trace likelihood shifts around the first complete calculation.

\subsection{Same-trace scoring}
\label{app:samescore}

For trace origin $o\in\{\mathrm{base},\mathrm{RLVR}\}$ and benchmark $d$, let
$C_{o,d}$ and $E_{o,d}$ be base minus RLVR per-token log-likelihood
differences in the early and execution segments. We report
\begin{equation}\label{eq:did}
\mathrm{DiD}_{d}
=
(C_{\mathrm{base},d}-E_{\mathrm{base},d})
-
(C_{\mathrm{RLVR},d}-E_{\mathrm{RLVR},d}).
\end{equation}
On Qwen2.5-7B every estimate is positive and all six intervals lie above zero
(Table~\ref{tab:samescore}). The relative shift between base and RLVR is
larger before the first complete calculation than during execution, whichever
policy produced the scored trace.

\begin{table}[ht]
\centering
\caption{Origin-stratified difference in differences for Qwen2.5-7B, in nats
per token, with 95\% problem-cluster bootstrap intervals, 10{,}000 draws,
seed 0.}
\label{tab:samescore}
\footnotesize
\begin{tabularx}{\linewidth}{lYYrr}
\toprule
Benchmark & DiD & 95\% CI & Problems & Traces \\
\midrule
AIME24 & 0.205 & $[0.132,0.304]$  & 9   & 129 \\
AMC23  & 0.307 & $[0.264,0.352]$  & 36  & 484 \\
GSM8K  & 0.313 & $[0.302,0.325]$  & 491 & 7736 \\
MATH500 & 0.337 & $[0.320,0.355]$ & 460 & 6018 \\
Minerva & 0.193 & $[0.171,0.216]$ & 132 & 1499 \\
OlympiadBench & 0.273 & $[0.254,0.292]$ & 322 & 4075 \\
\bottomrule
\end{tabularx}
\end{table}

\subsection{Semantic boundary resegmentation}
\label{app:semanticaudit}

The first-calculation boundary in \S\ref{sec:beyond} is extracted by a
deterministic rule parser. We independently resegment the same trace pool
using blinded model-based annotations of the earliest complete numeric
calculation: the annotator receives only the prompt, response, and numbered
text units, without model identity, trace origin, or parser output, and a
deterministic resolver maps the returned span to token offsets. The resolver
succeeds on 19{,}160 of the 19{,}941 scored traces (96.1\%), and failure
rates differ by at most 1.85 percentage points between Base- and RLVR-origin
traces across benchmarks.

Using the same per-token NLL caches, we resegment each resolved trace at its
semantic boundary and measure, for each trace origin, the Base-minus-SimpleRL
per-token log-likelihood difference in the early segment minus the same
difference in the execution segment (Table~\ref{tab:semantic_resegmentation}).
Contrasting the two origins recovers the difference in differences of
\S\ref{app:samescore}, and the origin contrast is preserved on every
benchmark: Base-origin and RLVR-origin estimates have opposite signs with
problem-cluster intervals excluding zero. The localization is therefore not
an artifact of the rule-based boundary parser. Canonicalizing the annotated
spans likewise preserves the corpus-level diversity gap, with
first-calculation vocabulary entropy 7.438 versus 7.050 on GSM8K and 7.412
versus 7.000 on MATH500 (Base versus SimpleRL).

\begin{table}[ht]
\centering
\caption{Semantic-boundary early-minus-execution NLL shift for Qwen2.5-7B
Base versus SimpleRL, stratified by the policy that generated the scored
trace. Intervals use 10{,}000 problem-cluster bootstrap draws.}
\label{tab:semantic_resegmentation}
\setlength{\tabcolsep}{3pt}
\scriptsize
\begin{tabularx}{\linewidth}{lYYYY}
\toprule
Benchmark & Base-origin shift & RLVR-origin shift & Base traces & RLVR traces \\
\midrule
AIME24        & $-0.083$ $[-0.111,-0.056]$ & $+0.055$ $[0.039,0.072]$ & 58   & 71 \\
AMC23         & $-0.142$ $[-0.178,-0.109]$ & $+0.060$ $[0.050,0.069]$ & 230  & 242 \\
GSM8K         & $-0.226$ $[-0.240,-0.212]$ & $+0.114$ $[0.110,0.117]$ & 5133 & 2500 \\
MATH500       & $-0.180$ $[-0.194,-0.166]$ & $+0.064$ $[0.060,0.068]$ & 3404 & 2306 \\
Minerva       & $-0.079$ $[-0.096,-0.064]$ & $+0.055$ $[0.050,0.061]$ & 689  & 690 \\
OlympiadBench & $-0.113$ $[-0.123,-0.103]$ & $+0.062$ $[0.058,0.066]$ & 2077 & 1760 \\
\bottomrule
\end{tabularx}
\end{table}

A local linear regression in a $\pm20$-token window around the semantic
boundary attributes the shift to a broad early phase rather than a
single-token jump, so we use the first calculation as a semantic alignment
point, not as a claimed discontinuity.

\subsection{Sampling depth at 256 samples}
\label{app:n256}

Re-evaluating the Qwen2.5-7B pair on GSM8K and MATH500 with 256 samples over
200 problems per benchmark, large-budget accuracy nears saturation for both
policies while first-calculation and trace diversity stay several-fold lower
under SimpleRL (Table~\ref{tab:n256}).

\begin{table}[ht]
\centering
\caption{Sampling depth at 256 samples per problem, 200 problems per
benchmark.}
\label{tab:n256}
\scriptsize
\setlength{\tabcolsep}{4pt}
\begin{tabularx}{\linewidth}{llYYYYY}
\toprule
Model & Benchmark & \passk{64} & \passk{256} & Distinct first calcs. & First-calc.\ $H$ & Distinct traces \\
\midrule
Base 7B     & GSM8K & 0.993 & 1.000 & 8.98 & 1.054 & 148.5 \\
SimpleRL 7B & GSM8K & 0.973 & 0.975 & 2.20 & 0.247 & 22.0 \\
\midrule
Base 7B     & MATH500 & 0.950 & 0.965 & 10.20 & 0.829 & 158.9 \\
SimpleRL 7B & MATH500 & 0.937 & 0.950 & 3.32  & 0.288 & 25.2 \\
\bottomrule
\end{tabularx}
\end{table}

\subsection{Conditional recovery over longer horizons}
\label{app:handoff}

For standard math tasks we construct an analogue of an entrance family from
calculations that the base policy reaches in free sampling. From 64 base
rollouts per problem we retain up to the two most frequent first-calculation
families and measure the RLVR policy after supplying that calculation.
Table~\ref{tab:base_reachable} separates families that the RLVR policy does
and does not reach in its own 64-sample free rollouts.

\begin{table}[ht]
\centering
\caption{Conditional execution from base-reachable first-calculation families.
Intervals use problem-cluster bootstrap.}
\label{tab:base_reachable}
\small
\begin{tabularx}{\linewidth}{lYYYY}
\toprule
Benchmark & Families & RLVR-unreached & Overall completion & Unreached completion \\
\midrule
GSM8K   & 384 & 224 & 0.863 $[0.823,0.901]$ & 0.850 $[0.801,0.898]$ \\
MATH500 & 337 & 155 & 0.715 $[0.661,0.768]$ & 0.713 $[0.645,0.780]$ \\
\bottomrule
\end{tabularx}
\end{table}

Longer horizons nevertheless introduce an execution constraint. On 534
four-number \task problems we supply states at four successive depths and
compare steps 50 and 275 with 64 continuations per state. The depth-by-late
interaction is $\gamma=-0.0152$ with 95\% problem-cluster bootstrap interval
$[-0.0224,-0.0080]$: the late-policy completion advantage decreases as more
downstream computation remains. Conditional recovery and downstream execution
therefore coexist as distinct constraints on long trajectories.

\section{Scale, Pipelines, and Supervised Controls}
\label{app:posttraining}

The two \task runs identify the same regime under PPO and GRPO. This appendix
asks whether early concentration recurs at larger scale and whether higher
accuracy requires the same loss of first-calculation breadth.

\subsection{Scale comparison}
\label{app:scale}

The 7B and 14B Qwen2.5 base and SimpleRL pairs move in the same direction: at
both scales RLVR raises single-sample accuracy while first-calculation entropy
falls sharply, and the matched GSM8K correct-only comparison agrees after
conditioning on correctness (Table~\ref{tab:scale}). First-calculation entropy
is the primary comparison because longer traces do not inflate it; the
distinct-trace rate in Table~\ref{tab:posttraining} is its length-sensitive
complement.

\begin{table}[ht]
\centering
\caption{Scale comparison for the Qwen base and SimpleRL pair. Macro columns
average six benchmarks; the last column is matched GSM8K correct-only
first-calculation entropy.}
\label{tab:scale}
\small
\begin{tabularx}{\linewidth}{lYYYY}
\toprule
Model & \passk{1} & \passk{64} & First-calc.\ $H$ & GSM8K correct-only $H$ \\
\midrule
Qwen Base 7B      & 0.379 & 0.754 & 1.063 & 0.894 \\
Qwen SimpleRL 7B  & 0.510 & 0.760 & 0.744 & 0.285 \\
\midrule
Qwen Base 14B     & 0.377 & 0.774 & 1.207 & 0.970 \\
Qwen SimpleRL 14B & 0.546 & 0.774 & 0.711 & 0.300 \\
\bottomrule
\end{tabularx}
\end{table}

\subsection{RL after distillation}
\label{app:deepscaler}

RL applied to a distilled 1.5B policy gives a different profile: single-sample
accuracy rises while large-budget accuracy and first-calculation structure
stay level, and the small entropy changes point in opposite directions on
GSM8K and MATH500 (Table~\ref{tab:deepscaler}). With the OLMo-3 comparison in
Table~\ref{tab:posttraining}, accuracy gains arrive with distinctly different
breadth profiles.

\begin{table}[ht]
\centering
\caption{RL after distillation at 1.5B; first-calculation statistics are
measured at the first complete calculation, 1{,}000 bootstrap draws.}
\label{tab:deepscaler}
\footnotesize
\begin{tabularx}{\linewidth}{llYYYY}
\toprule
Model & Benchmark & \passk{1} & \passk{64} & First-calc.\ $H$ & Effective first calcs. \\
\midrule
Distilled base 1.5B  & GSM8K & 0.701 & 0.985 & 1.126 & 3.67 \\
RL over distill 1.5B & GSM8K & 0.766 & 0.980 & 1.206 & 4.02 \\
\midrule
Distilled base 1.5B  & MATH500 & 0.633 & 0.955 & 1.168 & 4.26 \\
RL over distill 1.5B & MATH500 & 0.732 & 0.955 & 1.130 & 3.93 \\
\bottomrule
\end{tabularx}
\end{table}

\subsection{Supervised fine-tuning controls on \task}
\label{app:sft}

To assess whether solution narrowing is an inevitable consequence of learning
correct reasoning paths, we train supervised fine-tuning (SFT) baselines
starting from the exact base model of the GRPO series (Qwen2.5-3B-Instruct).

\paragraph{Disjoint multi-solution supervision.}
We construct 8{,}000 training problems and 500 validation problems disjoint
from the 135 held-out evaluation tasks by semantic key ($\text{sorted
numbers} + \text{target}$). For each training problem, a solver enumerates
all valid canonical expressions. We extract up to $k$ solutions per problem
using an entrance-diverse round-robin strategy across opening families and
render them into native \task reasoning traces using 12 templated prompt
variations. All generated traces undergo automated mathematical verification,
canonical parser validation, and deduplication.

\paragraph{Pre-registered checkpoint selection.}
Models are fine-tuned using LoRA (rank 32, $\alpha=64$, learning rate
$10^{-4}$, 2 epochs, batch size 16), with checkpoints saved every 250
optimizer steps. Checkpoint selection is performed on the 500 independent
validation tasks using the pre-registered rule
\begin{equation}
\hat{c} = \arg\min_{c:\ \mathrm{format}(c)\ge 0.90}\ \bigl|\widehat{\mathrm{pass@1}}_{\mathrm{val}}(c) - 0.429\bigr|,
\end{equation}
which selects the checkpoint closest to the GRPO step-450 operating point
($0.429$) subject to native-format validity above 0.90. The 135 evaluation
problems are then evaluated with 320 samples only at the selected checkpoint;
Table~\ref{tab:sft_k_ablation} reports the resulting SFT operating points.

\begin{table}[ht]
\centering
\caption{Dose--response ablation over supervised solutions per problem $k$ on
135 held-out \task tasks (320 samples per problem). All intervals are 2{,}000
problem-bootstrap draws (seed 1729).}
\label{tab:sft_k_ablation}
\small
\begin{tabularx}{\linewidth}{lYYYYY}
\toprule
Supervision condition & \passk{1} & \passk{64} & \passk{256} & Solution cov. & Entrance cov. \\
\midrule
GRPO step 450 (RLVR)  & 0.429 & 0.570 & 0.621 & 0.268 & 0.571 \\
\midrule
SFT ($k{=}1$)         & 0.120 $[0.095, 0.145]$ & 0.737 & 0.856 & 0.471 $[0.412, 0.532]$ & 0.604 \\
SFT ($k{=}2$)         & 0.221 $[0.176, 0.266]$ & 0.806 & 0.893 & 0.551 $[0.494, 0.609]$ & 0.766 \\
SFT ($k{=}4$, 8k)     & 0.310 $[0.254, 0.366]$ & 0.853 & 0.923 & 0.725 $[0.669, 0.782]$ & 0.825 \\
SFT ($k{=}4$, 16k)    & 0.333 $[0.274, 0.389]$ & 0.855 & 0.916 & 0.709 $[0.650, 0.764]$ & 0.855 \\
SFT ($k{=}4$, uniform)& 0.269 $[0.219, 0.315]$ & 0.836 & 0.928 & 0.779 $[0.725, 0.835]$ & 0.821 \\
SFT ($k{=}8$)         & 0.293 $[0.241, 0.344]$ & 0.836 & 0.903 & 0.766 $[0.710, 0.822]$ & 0.818 \\
\bottomrule
\end{tabularx}
\end{table}

\paragraph{Dose--response relationship.}
As shown in Table~\ref{tab:sft_k_ablation}, solution coverage exhibits a
strict monotonic increase with supervision breadth $k$ ($0.471 \to 0.551 \to
0.709 \to 0.766$). Even with a single supervised path per problem ($k{=}1$),
SFT preserves 0.471 coverage, exceeding the late RL policy (0.268). This
comparative demonstration indicates that solution space contraction is not an
unavoidable byproduct of learning correct reasoning paths, but reflects the
selective reinforcement dynamics of on-policy RL.

\paragraph{Continuation under GRPO.}
Starting from the $k{=}4$ SFT checkpoint (coverage $0.709$), we continue
on-policy GRPO training for 500 steps under the native \task reward. Across
the final 100 training updates, rollout response entropy reaches a mean of
$0.085$~nats: the concentration dynamic re-emerges even from a high-diversity
supervised initialization.

\section{Raw Trajectories for Qualitative Case Studies}
\label{app:casestudy}
This section reproduces the unedited verbatim model completions for the case studies evaluated at $T=0.6, p=0.95$ discussed in \S\ref{sec:beyond}.

\subsection{GSM8K Item 108 (Baking Cookies, Gold Answer: 50)}

\paragraph{Qwen Base 7B --- Reverse Execution Route (Sample Index 6; Correct):}
\begin{quote}
\scriptsize\ttfamily
To find out how many cookies Henry baked last year, we can follow these steps:
\begin{enumerate}[leftmargin=*,nosep]
\item \textbf{Determine the number of cookies Henry meant to bake this year:}
Henry now has 110 cookies. He dropped 5 cookies, so before dropping them, he had 110 + 5 = 115 cookies. He baked 15 more cookies than he meant to, so the number of cookies he meant to bake is 115 - 15 = 100 cookies.
\item \textbf{Calculate the number of cookies Henry baked last year:}
Henry wants to make twice as many cookies as last year. This year, he meant to bake 100 cookies, which is twice the number he baked last year. Therefore, the number of cookies he baked last year is 100 / 2 = 50 cookies.
\end{enumerate}
So, the number of cookies Henry baked last year is \boxed{50}.
\end{quote}

\paragraph{Qwen SimpleRL 7B --- Collapsed Algebraic Route (Sample Index 0; Correct):}
\begin{quote}
\scriptsize\ttfamily
To determine how many cookies Henry baked last year, we will follow these steps:
\begin{enumerate}[leftmargin=*,nosep]
\item \textbf{Define the variable:} Let $x$ be the number of cookies Henry baked last year.
\item \textbf{Calculate the number intended:} He intended to bake $2x$ cookies this year.
\item \textbf{Account for extra baked:} He baked $2x + 15$ cookies.
\item \textbf{Account for cookies dropped:} After dropping 5, he has $(2x + 15) - 5$.
\item \textbf{Set up the equation:} $(2x + 15) - 5 = 110 \implies 2x + 10 = 110 \implies 2x = 100 \implies x = 50$.
\end{enumerate}
So, Henry baked \boxed{50} cookies last year.
\end{quote}

\paragraph{OLMo-3 7B RLVR --- Preserved Reverse Route (Sample Index 12; Correct):}
\begin{quote}
\scriptsize\ttfamily
Let $x$ be the number of cookies Henry baked last year. Henry wants to make twice as many cookies as last year this year, so his intended amount is $2x$. When he finishes baking, he baked 15 more cookies than he meant to: $2x + 15$. Next, Henry drops 5 cookies, and then has 110 left. So before dropping the 5 cookies, he had: $110 + 5 = 115\text{ cookies}$. This 115 cookies is the actual number he baked: $2x + 15 = 115 \implies 2x = 115 - 15 = 100 \implies x = 100 / 2 = 50$. So, Henry baked \boxed{50} cookies last year.
\end{quote}

\subsection{GSM8K Item 392 (Streaming Bundle, Gold Answer: 34)}

\paragraph{Qwen SimpleRL 7B --- Collapsed Initial Bundle Sum Route (Sample Index 4; Correct):}
\begin{quote}
\scriptsize\ttfamily
Total cost of Hulu and Disney Plus before discount: 10 + 10 = 20. With a 20\% discount, the bundle costs $20 \times (1 - 0.20) = 16$. The total cost of all streaming services is $16 + 10 = 26$. Tim saves $60 - 26 = \boxed{34}$ dollars per month.
\end{quote}

\paragraph{OLMo-3 7B RLVR --- Preserved Percentage Savings Route (Sample Index 19; Correct):}
\begin{quote}
\scriptsize\ttfamily
Hulu and Disney Plus normally cost \$10 each, so \$20 combined. Bundling saves 20\%, which is $20 \times 0.20 = \$4$ saved on the bundle. The cost of the bundle is $20 - 4 = \$16$. Netflix is \$10, so total streaming is $16 + 10 = \$26$. Cancelling the \$60 cable package saves $60 - 26 = \boxed{34}$.
\end{quote}

\subsection{Companion Analysis on MATH500 (Item 451)}
\label{app:math500_companion}
On MATH500 Item 451 (map scale ratio; gold answer $639/40$), all models achieve $\passk{64}=1$. Qwen Base accesses 10 distinct correct first-step forms ($H=1.590$~nats, $\passk{1}=0.797$). Direct RLVR (Qwen SimpleRL) elevates $\passk{1}$ to $1.000$ while collapsing the correct support to only 2 forms ($H=0.525$~nats, with 50/64 rollouts entering via $15+0.975$). In contrast, the staged OLMo-3 progression preserves opening diversity ($7 \to 10 \to 13$ correct forms; $H: 1.678 \to 1.696 \to 1.804$~nats across SFT, DPO, and RLVR), confirming that multi-stage entropy preservation generalizes across distinct mathematical reasoning benchmarks.

\section{Training and Evaluation Details}
\label{app:environment}

\paragraph{Hardware and software.}
All PPO training and all evaluations ran on a Linux server with Ubuntu 24.04
LTS, two Intel Xeon Gold 6330 CPUs, 503 GiB RAM, and two NVIDIA GeForce RTX
4090 GPUs. The stack is Python 3.10.20, PyTorch 2.4.0 with CUDA 12.1,
Transformers 4.57.1, vLLM 0.6.3, Ray 2.55.1, and VERL 0.1.

\paragraph{PPO training configuration.}
The Qwen2.5-3B \task run follows the TinyZero setup
\citep{pan2025tinyzero} with PPO. The actor uses AdamW with constant learning
rate $10^{-6}$, 160 prompts per iteration, mini-batch size 64, and micro-batch
size 4; the critic learning rate is $10^{-5}$. Rollouts use temperature 1.0,
top-$p=1.0$, and a 1{,}024-token response limit, while \task evaluation uses a
256-token limit. GAE uses $\gamma=1.0$ and $\lambda=1.0$, rollout group size is
$G=1$, and the actor KL loss is disabled. Training runs 15 epochs and 276
optimizer steps indexed 0 to 275, with seed 1 and checkpoints every 25 steps.

\section{Extended Related Work}
\label{app:related}

\paragraph{RLVR, empirical support, and reasoning boundaries.}
Modern RLVR systems build on outcome-based optimization in the GRPO and PPO
families
\citep{shao2024deepseekmath,deepseekai2025deepseekr1,yu2025dapo,zeng2025simplerlzoo}.
Large-$k$ evaluations identify regimes where RL improves sampling efficiency
without preserving the base model's empirical answer support
\citep{yue2025doesreinforcementlearningreally,wu2025invisibleleash}, while
process-aware evaluation and prolonged exploratory RL report expansion in
others
\citep{wen2025reinforcementlearningverifiablerewards,liu2025prorlprolongedreinforcementlearning}.
Further analyses tie contraction to amplification of pretraining-favored
modes, winner-take-all dynamics, and finite on-policy exposure
\citep{zhao2025echochamberrlposttraining,nguyen2025reasoningboundary,
yuan2026overtraining,zhou2026whenrlvrshrinks}. These works ask which prompts
or answers remain reachable. We ask where probability goes inside a prompt
that stays solvable.

\paragraph{Diversity-preserving optimization.}
Training-time methods counter sharpening by reweighting unlikely correct
rollouts, smoothing updates, changing divergence geometry, or reallocating
exploration \citep{he2025rewardingunlikelyliftinggrpo,
gai2025differentialsmoothingmitigatessharpening,
li2025choicedivergenceneglectedkey,
song2025outcomebasedexplorationllmreasoning,
yao2025diversityawarepolicyoptimizationlarge,
hu2026diversityincentivizedexplorationversatilereasoning}.
Our entrance-level analysis supplies a concrete target for them: probability
mass over alternative early solution families.

\paragraph{Forking decisions and prefix interventions.}
Token-level work identifies sparse decisions with outsized downstream
influence \citep{bigelow2024forkingpathsneuraltext,
wang20258020rulehighentropyminority,
kim2026rolloutsbegin}, and BODHI measures semantic branching from sampled
traces \citep{saha2026bodhi}. Prefix methods move the states from which
rollouts begin \citep{zhang2025bread,macar2026thoughtbranches}. Entrance
families differ in two ways: they partition the complete solver-enumerated
valid solution set, including families absent from model samples, and their
entrances can be supplied without copying a successful trace.

\paragraph{Inference-time breadth and checkpoint composition.}
Repeated sampling, self-consistency, Best-of-$N$, and verifier-guided search
exploit only modes that remain reachable under the sampling policy
\citep{brown2024largelanguagemonkeysscaling,
wang2023selfconsistencyimproveschainthought,
huang2025bestofnbestthemcoverage,
snell2024scalingtesttime}. Sampling across training checkpoints and
weight-space interpolation recover modes that fade during post-training
\citep{li2026temporalsampling,li2026attributing,dang2025weightensembling}. Our results tie those
methods to a location: they work when they restore diversity at the states
where solution families are chosen.

\paragraph{Measuring reasoning breadth.}
Recent metrics separate process correctness, breadth, depth, stability, and
semantic path diversity \citep{wen2025reinforcementlearningverifiablerewards,
dragoi2025passkbreadthdepthmetricsreasoning,
liu2025llmscapablestablereasoning,
ju2025reasoningpathdivergencenew,
yu2025passkdiagnostic}. On \task, exhaustive enumeration supplies a complete
denominator, the canonical solution set with its opening families, which is
what makes failure to enter a family distinguishable from failure to complete
it once entered.

\end{document}